\documentclass[twoside]{article}

\usepackage[accepted]{aistats2021}
\usepackage[utf8]{inputenc} 
\usepackage[T1]{fontenc}    
\usepackage{hyperref}       
\usepackage{url}            
\usepackage{nicefrac}       
\usepackage{microtype}      
\usepackage{subfigure}
\usepackage{booktabs} 
\usepackage{times}
\usepackage{amssymb,amsmath,amsfonts}
\usepackage{graphicx}   
\usepackage{verbatim}   
\usepackage{color}      
\usepackage{stmaryrd}
\usepackage{bbm}
\usepackage{multirow}
\usepackage{pifont}
\usepackage{textcomp}
\usepackage{dsfont}
\usepackage{graphicx}
\usepackage{algorithm}
\usepackage{algorithmic}

\usepackage[round]{natbib}

\usepackage{minitoc}

\newtheorem{theorem}{Theorem}
\newtheorem{definition}{Definition}
\newtheorem{lemma}{Lemma}

\newtheorem{remark}{Remark}

\newcommand{\bp}{\noindent{\emph{Proof}.}\ }
\newcommand{\ep}{\hfill $\Box$}

\newcommand{\BEAS}{\begin{eqnarray*}}
\newcommand{\EEAS}{\end{eqnarray*}}
\newcommand{\BEA}{\begin{eqnarray}}
\newcommand{\EEA}{\end{eqnarray}}
\newcommand{\BEQ}{\begin{equation}}
\newcommand{\EEQ}{\end{equation}}
\newcommand{\BIT}{\begin{itemize}}
\newcommand{\EIT}{\end{itemize}}
\newcommand{\BNUM}{\begin{enumerate}}
\newcommand{\ENUM}{\end{enumerate}}
\newcommand{\beq}{\begin{equation}}
\newcommand{\eeq}{\end{equation}}
\newcommand{\beqa}{\begin{eqnarray}}
\newcommand{\eeqa}{\end{eqnarray}}
\newcommand{\beqan}{\begin{eqnarray*}}
\newcommand{\eeqan}{\end{eqnarray*}}
\newcommand{\bealn}{\begin{align*}}
\newcommand{\eealn}{\end{align*}}
\newcommand{\al}[1]{ \begin{align} #1  \end{align}}

\newcommand{\als}[1]{ \begin{align*} #1  \end{align*}}

\newcommand{\sk}{\nonumber\\}

\newcommand{\el}{\end{flushleft}}
\newcommand{\bl}{\begin{flushleft}}

\newcommand{\argmin}{\arg\!\min}
\newcommand{\argmax}{\arg\!\max}
\newcommand{\Argmax}{\mathop{\mathrm{Argmax}}}

\newcommand{\bI}{\mathbb{I}}

\newcommand{\bS}{\mathbb{S}}

\newcommand{\cA}{\mathcal{A}}
\newcommand{\cB}{\mathcal{B}}
\newcommand{\cC}{\mathcal{C}}
\newcommand{\cD}{\mathcal{D}}
\newcommand{\cE}{\mathcal{E}}
\newcommand{\cF}{\mathcal{F}}
\newcommand{\cG}{\mathcal{G}}

\newcommand{\cK}{\mathcal{K}}
\newcommand{\cL}{\mathcal{L}}
\newcommand{\cM}{\mathcal{M}}

\newcommand{\cO}{\mathcal{O}}
\newcommand{\cP}{\mathcal{P}}

\newcommand{\cR}{\mathcal{R}}
\newcommand{\cS}{\mathcal{S}}
\newcommand{\cT}{\mathcal{T}}

\newcommand{\cV}{\mathcal{V}}

\newcommand{\cX}{\mathcal{X}}

\newcommand{\kR}{\mathfrak{R}}

\newcommand{\Nat}{\mathbb{N}}

\newcommand{\indic}[1]{\mathbb{I}\{#1\}}

\renewcommand{\phi}{\varphi}
\renewcommand{\epsilon}{\varepsilon}

\def\NN{{\mathbb N}}
\def\EE{{\mathbb E}}
\def\PP{{\mathbb P}}
\def\RR{{\mathbb R}}

\renewcommand{\Pr}{\mathbb{P}}
\newcommand{\supp}{\mathrm{supp}}

\usepackage[colorinlistoftodos, textwidth=4cm, shadow]{todonotes}

\newcommand{\vertiii}[1]{{\left\vert\kern-0.25ex\left\vert\kern-0.25ex\left\vert #1
    \right\vert\kern-0.25ex\right\vert\kern-0.25ex\right\vert}}

\newcommand{\AlgB}{{\textcolor{red!50!black}{\texttt{DBN-UCRL}}}}
\newcommand{\EVI}{\texttt{EVI}}

\begin{document}

%

%

\twocolumn[

\aistatstitle{Improved Exploration in Factored Average-Reward MDPs}

\aistatsauthor{Mohammad Sadegh Talebi \And Anders Jonsson \And  Odalric-Ambrym Maillard }

\aistatsaddress{Department of Computer Science \\University of Copenhagen \And ICT Department\\ Universitat Pompeu Fabra \And Univ.~Lille, Inria, CNRS, Centrale Lille \\ UMR 9189 -- CRIStAL, F-59000 Lille, France } 
]

\begin{abstract}
We consider a regret minimization task under the average-reward criterion in an unknown Factored Markov Decision Process (FMDP). More specifically, we consider an FMDP where the state-action space $\mathcal X$ and the state-space $\mathcal S$ admit the respective factored forms of $\mathcal X = \otimes_{i=1}^n \mathcal X_i$ and $\mathcal S=\otimes_{i=1}^m \mathcal S_i$, and the transition and reward functions are factored over $\mathcal X$ and $\mathcal S$. Assuming known factorization structure, we introduce a novel regret minimization strategy inspired by the popular UCRL2 strategy, called \AlgB, which relies on Bernstein-type confidence sets defined for individual elements of the transition function. We show that for a generic factorization structure, \AlgB\ achieves a regret bound,  whose leading term strictly improves over existing regret bounds in terms of the dependencies on the size of $\cS_i$'s and the involved diameter-related terms. We further show that when the factorization structure corresponds to the Cartesian product of some base MDPs, the regret of \AlgB\ is upper bounded by the sum of regret of the base MDPs. We demonstrate, through numerical experiments on standard environments, that \AlgB\ enjoys a substantially improved regret empirically over existing algorithms that have frequentist regret guarantees. 
\end{abstract}

\vspace{-2mm}
\section{INTRODUCTION}\label{sec:intro}
\vspace{-2mm}
In reinforcement learning (RL), an agent repeatedly interacts with an unknown environment in order to maximize its cumulative reward. A typical model of the environment is a Markov decision process (MDP): In each time step, the agent observes a state, takes an action and receives a reward before transiting to the next state. To achieve its objective, the agent has to estimate the parameters of the MDP from experience and learn a policy that maps states to actions. While doing so, the agent faces a choice between two basic strategies: \emph{Exploration}, i.e.~discovering the effects of actions on the environment, and \emph{exploitation}, i.e.~using its current knowledge to maximize reward in the short term.

Most model-based RL algorithms treat the state as a black box. In many practical cases, however, the environment exhibits structure that can be exploited to learn more efficiently. A common form of such structure is {\em factorization}. In a Factored MDP (FMDP), (see, e.g., \citet{boutilier1999fmdps}), the state-space $\cS=\otimes_{i=1}^m\cS_{i}$ and action space $\cA=\otimes_{i=1}^{n-m} \cA_i$ are composed of $m$ and $n-m$ individual factors, respectively. In this context, a state-action pair $x=(s,a)\in \cX:= \cS\times\cA$ is a tuple of $n$ factor values. Each state factor $\cS_i$ has its own transition function $P_i$, and the new factor value of $\cS_i$, as a result of applying action $a$ in state $s$, only depends on a small subset of the factors in $\cS\times\cA$. For $\cS_i$, the set
$Z_i\subset\{1,\ldots,n\}$,  termed the \emph{scope} of $\cS_i$, collects the indices of relevant factors for $\cS_i$. Then $P_i$ only depends on $\cX[Z_i]:=\otimes_{i\in Z_i}\cX_i\subset \cX$. Namely, $\cS_i$ is conditionally independent of factors with indices outside $Z_i$. This \emph{conditional independence structure} can be exploited to compactly represent the parameters of an FMDP.
(We present a complete definition of FMDPs in Section \ref{sec:model}.) 

In this paper we consider the problem of regret minimization in FMDPs. Regret measures how much more reward the agent could have obtained using the best stationary policy, compared to the actual reward obtained. To achieve low regret, the agent must carefully balance exploration and exploitation: An agent that explores too much will not accumulate enough reward, while the one exploiting too much may fail to discover high-reward regions of the state-space. 

\vspace{-2mm}
\paragraph{Related Work.}Factored state representations have been used since the early days of artificial intelligence \citep{fikes1971strips}. In RL, factored states were first proposed as part of Probabilistic STRIPS \citep{boutilier1994fmdps}. When the FMDP structure and parameters are known, researchers have proposed two main approaches for efficiently learning a policy. The first approach consists in maintaining and updating a structured representation of the policy (\citet{boutilier1999fmdps,poupart2002fmdps,degris2006fmdps,raghavan2015fmdps}), whereas the second is to perform linear function approximation over a set of basis functions (\citet{guestrin2003fmdps,dolgov2006fmdps,szita2008fmdps}). However, only a few theoretical guarantees for these approaches exist. When the FMDP structure and parameters are unknown, several authors have proposed algorithms for structure learning (\citet{kearns1999fmdps,strehl2007fmdps,diuk2009fmdps,chakraborty2011fmdps,hallak2015fmdps,guo2018sample,rosenberg2020oracle}). Many of these algorithms admit PAC-type guarantees on their sample complexities.

The focus of this paper is RL in an FMDP under the average-reward criterion, in an intermediate setting where the underlying structure of the FDMP is \textit{known}, while actual reward and transition distributions are \textit{unknown}. There is a rich and growing literature on average-reward RL in finite non-factored MDPs, where several algorithms with theoretical regret guarantees are presented 
 (e.g.,  \citet{burnetas1997optimal,jaksch2010near,bartlett2009regal,fruit2018efficient,talebi2018variance,zhang2019regret,qian2019exploration,
bourel2020tightening,wei2020model}) \footnote{Besides this growing line of research, some papers study RL in episodic MDPs; see, e.g., \citep{azar2017minimax,dann2017unifying}.}. 

Despite such a rich literature in non-factored MDPs, RL in FMDPs has received relatively less attention, and only a few algorithms with performance guarantees in terms of regret or sample complexity are known. Among a few existing works, \cite{kearns1999efficient,szita2009optimistic} study RL in discounted FMDPs presenting DBN-E$^3$ and FOIM, respectively. 
In the regret setting, \cite{osband2014near} present the first algorithms with provably sublinear regret, and are followed very recently by \cite{xu2020near,tian2020towards,chen2021efficient,rosenberg2020oracle}. Except \citep{rosenberg2020oracle}, all these works assume a known structure. 
In the episodic setting, \cite{osband2014near} present Factored-UCRL achieving a regret of $\widetilde \cO(D\sum_{i=1}^m\sqrt{S_i|\cX[Z_i]|T})$ after $T$ steps.\footnote{The notation $\widetilde O(.)$ hides poly-logarithmic terms in $T$.} (To simplify the presentation, in this section we assume that the reward and transition functions have the same scope sets.) Here, 
$D$ denotes the diameter of the FMDP (for a precise definition, see the footnote in Section \ref{sec:model}). \cite{tian2020towards} present two algorithms, F-EULER and F-UCBVI, which are extensions of UCBVI-CH \citep{azar2017minimax} and EULER \citep{zanette2019tighter} to FMDPs, respectively. In particular, F-EULER achieves a minimax-optimal regret of $\widetilde \cO(\sum_{i=1}^m \sqrt{H|\cX[Z_i]| T})$ for a rich class of structures, where $H$ denotes the fixed episode length. In the average-reward setting, \cite{xu2020near} present two oracle-efficient algorithms, DORL and PSRL, which admit efficient implementations when an efficient oracle exists. DORL achieves a regret of $\widetilde \cO(D\sum_{i=1}^m \sqrt{S_i|\cX[Z_i]| T})$. The main objective in \citep{xu2020near} is to design a computationally efficient algorithm (with sublinear regret), for when an efficient oracle exists. 
RL in FMDPs with unknown structure are seldom studied in the literature. To the best of our knowledge, \citep{rosenberg2020oracle} is the only work presenting an algorithm with provable regret in FMDPs without any prior knowledge of the structure. The presented algorithm, SLF-UCRL, combines the structure learning method of \citep{strehl2007fmdps} with  DORL \citep{xu2020near}. Thus, it is oracle-efficient, like DORL. 
In contrast to \citep{xu2020near} and \citep{rosenberg2020oracle}, we do not address the problem of efficient planning in FMDPs and instead aim for statistical efficiency from both theoretical and empirical standpoints. 

We finally mention that some papers, notably \citep{zimmert2018factored}, study regret minimization in factored bandit problems, where the action-space is a Cartesian product of some atomic sets. Following \cite{osband2014model}, recent literature on FMDPs (including the present paper) consider a factored action-space, which includes the Cartesian product as a special case. Nonetheless, the key feature that makes FMDPs suitable to model large decision problems is their factored dynamics. (In practice, the action-space may not be factored.) More importantly, the corresponding challenges of RL in FMDPs are due to the factored transition function, for which the technical tools developed for factored bandit problems could not be directly used.


\vspace{-2mm}
\paragraph{Outline and Contributions.}
We introduce in Section~\ref{sec:algo} \AlgB, a novel algorithm for average-reward RL in FMDPs, assuming a known factorization structure. 
\AlgB\ is a model-based algorithm maintaining confidence sets for transition and reward functions. Specifically, it maintains tight Bernstein-type confidence sets for $P_i,i=1,\ldots,m$, in contrast to $L_1$-type confidence sets used in DORL and UCRL-Factored. On the theoretical side, we derive
 finite-time regret upper bounds for \AlgB\ demonstrating the potential gain of using such confidence sets in terms of regret: For generic structures, we report a regret upper bound (in Theorem \ref{thm:regret_Alg1}) scaling as $\widetilde \cO\big(\sum\nolimits_{i=1}^m\!\sqrt{\sum\nolimits_{(s,a)\in\cX[Z_i]} D^2_{i,s}K_{i,s,a}T}\big)$, where $D_{i,s}$ is a notion of diameter termed \emph{factored diameter} (Definition \ref{def:factored_diameter}) and $K_{i,s,a}$ denotes the number of next-states for $P_i$ under $(s,a)$. \AlgB\ achieves a strictly smaller regret than existing ones: (i) In contrast to previous bounds that depend on the (global) diameter $D$ of the FMDP, this bound depends on the factored diameter which is tighter and problem-dependent; (ii) it improves the dependency of the regret on $S_i$ to $K_{i,s,a}$. The factored diameter is always smaller than $D$: There exist cases, as illustrated in Section \ref{sec:results}, where $D$ may scale as $S:=|\cS|$, whereas $D_{i,s}$ could scale as $\max_{a} K_{i,s,a}$. Hence, $D_{i,s}$ could be exponentially (in $m$) smaller than $D$. 
Our second result concerns specific structures in the form of Cartesian products of some base MDPs. 
Theorem \ref{thm:regret_Alg1_Cartesian} shows that in Cartesian products, \AlgB\ incurs the \emph{sum} of regret of each underlying base MDP. This latter result significantly improves over previous regret bounds for the product case that were unable to establish a fully localized regret bound. This includes the bounds of \citep{osband2014near} and \citep{xu2020near} that would still depend on the global diameter $D$ of the FMDP in this case. This leads to a term that is exponentially smaller (in $m$) than $D$. In Section~\ref{sec:xps}, through numerical experiments, we show that on standard environments \AlgB\ significantly outperforms other state-of-the-art algorithms that have frequentist regret guarantees.  


\vspace{-2mm}
\paragraph{Notations.}We introduce some notations that will be used throughout. 
Given sets $\cX$ and $\cS$, let $\cR_{\cX,[0,1]}$ be the set of all reward functions on $\cX$ with image bounded in $[0,1]$,
and let $\cP_{\cX,\cS}$ be the set of all transition functions from $\cX$ to $\cS$, i.e.~$P\in \cP_{\cX,\cS}$ satisfies: For all $x\in\cX$, $P(\cdot|x)$ is a probability distribution over $\cS$, i.e.~$P(s|x)\ge 0$ for all $s\in \cS$ and $\sum_{s\in\cS}P(s|x)=1$.
For a distribution $q$, $\supp(q)$ denotes the support set of $q$. For $n\in \NN$, let $[n]:=\{1,\ldots,n\}$. $\indic{\cdot}$ denotes the indicator function of an event.

\vspace{-2mm}\noindent
\section{PROBLEM FORMULATION}\label{sec:model}
\vspace{-2mm}\noindent
We study a learning task in a finite MDP $M = (\cS, \cA, P, R)$ under the average-reward criterion, where $\cS$ denotes the set of states with cardinality $S$, $\cA$ denotes the set of actions (available at each state) with cardinality $A$, and $P$ and $R$ denote the transition and reward functions, respectively. Choosing action $a\in \cA$ in state $s\in \cS$ results in a transition to a state $s'\sim P(\cdot|s,a)$ and a reward drawn from $R(s,a)$, with mean $\mu(s,a)$. 
We assume that $M$ is an FMDP, namely its transition and reward functions admit some conditional independence structure, as detailed below.  

\vspace{-2mm}\noindent
\paragraph{Factored Representations.}
To formally describe the factored structure, we introduce a few notations and definitions that are standard in the literature on FMDPs (see, e.g., \cite{szita2009optimistic,osband2014near}). We begin by introducing the \emph{scope} operator for a factored set. 

\begin{definition}[{Scope Operator for a Factored Set $\cX$}]
Let $\cX\!=\!\otimes_{i=1}^n \cX_i$ be a finite factored set. For any subset of indices $Z\!\subseteq \![n]$, we define $\cX[Z]\!:=\!\otimes_{i\in Z} \cX_i$. Moreover, for any $x\!\in\!\cX$, we let $x[Z]\!\in\!\cX[Z]$ denote the value of the variables $x_i\!\in\! \cX_i$ with indices $i\!\in\! Z$. For $i\!\in\! [n]$, we will write $x[i]$ as a shorthand for $x[\{i\}]$. 
\end{definition}

\vspace{-1mm}\noindent
An FMDP is represented by a tuple 
\mbox{$
M\!=\!\big(\{\cS_i\}_{i\in [m]},\!\{\cX_i\}_{i\in [n]},\!\{P_i\}_{i\in [m]},\!\{Z_i^p\}_{i\in [m]},\!\{R_i\}_{i\in [\ell]},$} $\{Z_i^r\}_{i\in [\ell]}\big)\, ,
$
where $\cS_i$ is the $i$-th state factor, $\cX_i$ is the $i$-th state-action factor, $P_i$ is the transition function associated with $\cS_i$, $R_i$ is the $i$-th reward function, and $Z_i^p$ (resp.~$Z_i^r$) denotes the scope set of $P_i$ (resp.~$R_i$) for $\cX\!=\!\otimes_{i=1}^n \cX_i$. The state-space is $\cS\!=\!\otimes_{i=1}^m\!\cS_i$ and the state-action space is $\cX\!=\!\cS_1\!\otimes\!\cdots\!\otimes\!\cS_m\!\otimes\!\cA_1\!\otimes\!\cdots\!\otimes\!\cA_{n-m}\!=\!\cS\times\cA$, where $\cA=\cA_1\otimes\cdots\otimes\cA_{n-m}$ is the (possibly factored) action-space.\footnote{Some authors have studied compact representations of the state-action space, such as decision trees, but we do not consider such representations in the present paper.}


\begin{definition}[{Factored Reward Functions}]\label{def:R_factored}
The class $\cR$ of reward functions is factored over $\cX\!=\!\otimes_{i=1}^n \cX_i$ with scopes $Z_1^r,\ldots,Z_\ell^r$ if and only if for all $R\!\in\! \cR$ and $x\!\in\! \cX$, there exist $\{R_i\!\in\! \cR_{\cX[Z_i^r],[0,1]}\}_{i\in [\ell]}$ such that 
any realization $r\sim R(x)$ implies $r=\sum_{i=1}^\ell r[i]$ with $r[i]\sim R_i(x[Z_i^r])$. Furthermore, let us define $r^{\text{col}}= \tfrac{1}{\ell}\sum_{i=1}^\ell r[i]$. 
\end{definition}

Without loss of generality, we assume that the rewards of each factor are bounded in $[0,1]$. Note that the collected reward $r^{\text{col}}$ is, by definition, bounded in $[0,1]$ too.

\begin{definition}[{Factored Transition Functions}]\label{def:P_factored}
The class $\cP$ of transition functions is factored over $\cX=\otimes_{i=1}^n \cX_i$ and $\cS = \otimes_{i=1}^m \cS_i$ with scopes $Z_1^p,\ldots,Z_m^p$ if and only if for all $P\in \cP$ and $x\in \cX$ and $s\in \cS$, there exist $\{P_i\in \cP_{\cX[Z_i^p], \cS_i}\}_{i\in [m]}$ such that 
$
P(s|x) = \prod_{i=1}^m P_i\big(s[i]\big|x[Z_i^p]\big) \, .
$
\end{definition}

In order to clarify the presentation of confidence sets in the subsequent sections, we further introduce the following more compact representation of an FMDP. Let $\cG_r = \big(\{\cX_i\}_{i\in [n]}; \{Z_i^r\}_{i\in [\ell]}\big)$ and $\cG_p = \big(\{\cX_i\}_{i\in [n]}, \{\cS_i\}_{i\in [m]}; \{Z_i^p\}_{i\in [m]}\big)$. We compactly represent an FMDP with structure $\cG= \cG_r \cup \cG_p$ by a tuple
$M = \big(\{P_i\}_{i\in [m]}, \{R_i\}_{i\in [\ell]}; \cG\big),$
and let $\cG(M)$  denotes its  corresponding structure. We finally introduce the set $\mathbb{M}_\cG$ of all FMDPs with structure  $\cG$: 
$$
\mathbb M_{\cG}\!=\!\Big\{M\!=\!(P,\!R;\!\cG)\!:\! P\!\in\!\cP_{\cX,\cS}^{\textrm{fac}}(\cG_p) \hbox{ and }  R\!\in\!\cR_{\cX,[0,1]}^{\textrm{fac}}(\cG_r)\Big\}\, ,
$$
where $\cP_{\cX,\cS}^{\textrm{fac}}(\cG_p)$ (resp.~$\cR_{\cX,[0,1]}^{\textrm{fac}}(\cG_r)$) denotes the set of transition (resp.~reward) functions satisfying Definition \ref{def:P_factored} (resp.~Definition \ref{def:R_factored}).

\begin{remark}
An FMDP $M$ can be represented by a Dynamic Bayesian Network (DBN) $\cB=(\cV,\cE,\cT)$, where $\cV$ is a set of $m$ discrete variables $\{v_i\}_{i\in[m]}$ and $\ell$ continuous variables $\{u_j\}_{j\in[\ell]}$, duplicated on two timeslices, $\cE$ is a set of edges between the two timeslices, and $\cT$ is a set of conditional probability tables (CPTs). In this case, $\cS_i = \cD(v_i)$ is the domain of variable $v_i$, $i\in[m]$, $\cX[Z_i^p]$ (resp.~$\cX[Z_j^r]$) are the elements used to index the rows of the CPT of variable $v_i$ (resp.~$u_j$), and $\cX[Z_i^p]$ (resp.~$\cX[Z_j^r]$) distinguishes between elements of $\cS_k$, $k\in[m]$, if and only if $(v_k,v_i)\in\cE$ (resp.~$(v_k,u_j)\in\cE$). In this context, $\cG$ is the structure of the DBN $\cB$ while $P$ and $R$ are the parameters of the CPTs in $\cT$. By a slight abuse of terminology, we refer to $\cG(M)$ as the DBN structure of the FMDP $M$. 
\end{remark}

\begin{figure}
	\centering
	\includegraphics[scale=.26]{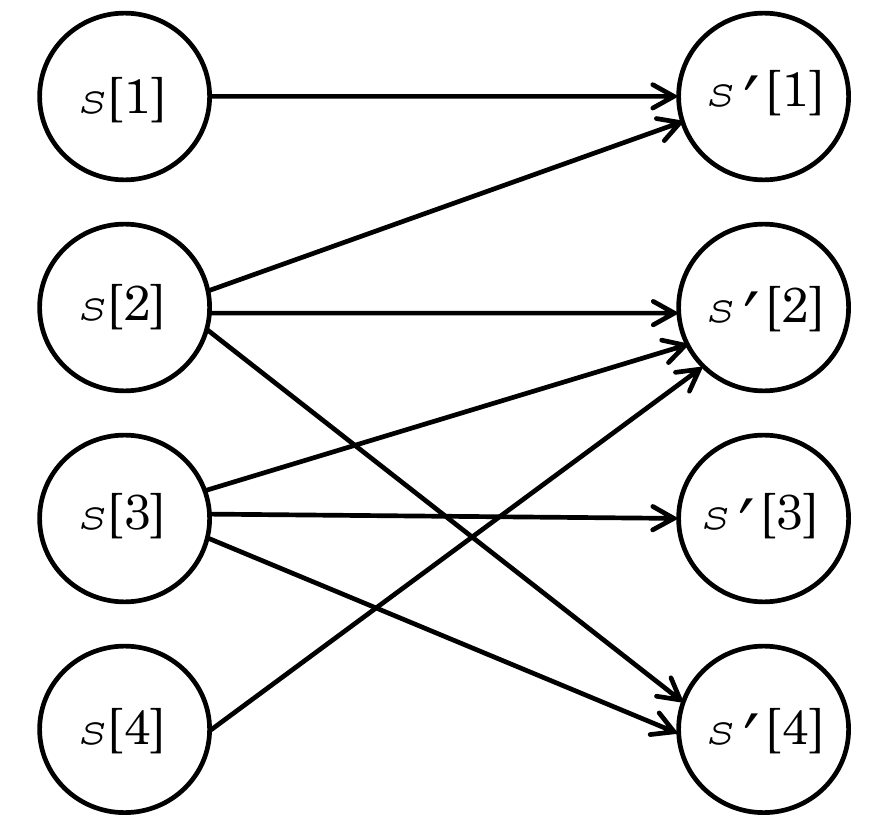}
	\caption{An example of a DBN characterizing the conditional independence structure of an FMDP.}
	\label{fig:FDMP_example}
\end{figure}

To help understand our notations,  we provide an example of an FMDP, whose conditional independence structure is represented using the DBN shown in Figure \ref{fig:FDMP_example}. The state-space has $m=4$ factors. For simplicity, we assume that all state factors are identical and equal to $\{a,b,c\}$ and that the action-space is non-factored. Nodes on the left-hand side of the DBN correspond to the current state $s$, whereas those on the right-hand side represent the next state $s'$, with each node representing a random variable corresponding to the value of a factor. For this DBN, the scopes are given by $Z_1^p=\{1,2\}$, $Z_2^p=\{2,3,4\}$, $Z_3^p=\{3\}$, and $Z_4^p=\{2,3\}$. Hence, for example, the resulting value of factor $\cS_1$ is independent of factors $\cS_3$ and $\cS_4$.
Furthermore, $\cX = \{a,b,c\}^4 \times \cA$, $\cX[Z_1^p] = \{a,b,c\}^2\times \cA$, $\cX[Z_2^p] = \{a,b,c\}^3\times \cA$, $\cX[Z_3^p] = \{a,b,c\}\times \cA$, and $\cX[Z_4^p] = \{a,b,c\}^2\times \cA$.

\paragraph{Regret Minimization in FMDPs.}
We consider a finite FMDP $M = \big(\{P_i\}_{i\in [m]}, \{R_i\}_{i\in [\ell]}; \cG\big)$ and the following RL task. An agent interacts with $M$ for $T$ rounds, starting in an initial state $s_1\in \cS$ chosen by Nature. At each time $t$, the agent is in state $s_t = (s_t[1],\ldots,s_t[m])$ and chooses an action $a_t$ based on its observations so far. Let $x_t=(s_t,a_t)$ denote the state-action pair of the agent at time $t$. Then, (i) it receives a reward vector $r_t = (r_t[1],\ldots, r_t[\ell])$, where for each $i\in [\ell]$, $r_t[i]\sim R_i(x_t[Z_i^r])$; and (ii) Nature decides a next state $s_{t+1} = (s_{t+1}[1],\ldots,s_{t+1}[m])$ where for each $i\in [m]$, $s_{t+1}[i]\sim P_i(\cdot|x_t[Z_i^p])$. Let $r^{\text{col}}_t = \tfrac{1}{\ell} \sum_{i\in [\ell]} r_t[i]$ denote the normalized collected reward at time $t$. 

The goal of the agent is to maximize the cumulative reward $\sum_{t=1}^T r^{\text{col}}_t = \sum_{t=1}^T \tfrac{1}{\ell} \sum_{i\in [\ell]} r_t[i]$, where $T$ denotes the time horizon. We assume that the agent perfectly knows $\cG$, but not the transition function $P$ or the  reward function $R$. It therefore has to learn them by trying different actions and recording the realized rewards and state transitions. The performance of the agent can be assessed resorting to the notion of regret. Following \cite{jaksch2010near}, we define the regret of a learning agent (or algorithm) $\mathbb A$, after $T$ steps and starting from an initial state $s_1\!\in\! \cS$, as:
\als{
\kR(T, \mathbb A, s_1) = Tg^\star(s_1) - \sum_{t=1}^T r^{\text{col}}_t \, ,
}
where $g^\star$ denotes the long-term average-reward (or gain) of $M$, in terms of $r^{\text{col}}$, starting from state $s_1$; we refer to \citep{puterman2014markov} for further details. Alternatively, the objective of the agent is to minimize the regret, which calls for balancing exploration and exploitation.
In this paper we consider communicating MDPs, for which the gain does not depend on $s_1$, that is, $g^\star(s_1) \!=\! g^\star$ for all $s_1\!\in\! \cS$. We therefore define: $\kR(T, \mathbb A) = Tg^\star \!- \!\sum_{t=1}^T  r^{\text{col}}_t$. The class of communicating MDPs arguably captures a big class of RL tasks of practical interest, and most literature on regret minimization in the average-reward setting has developed algorithms for this class. A notable property of communicating MDPs is having a finite {\it diameter}, as formalized in \citep{jaksch2010near}.\footnote{Given an MDP $M$, the diameter $D := D(M)$ is defined as $D(M) := \max_{s\neq s'} \min_{\pi:\cS\to \cA} \EE[T^\pi (s,s')]$, where $T^\pi (s,s')$
denotes the number of steps it takes to get to $s'$ starting from $s$ and following policy $\pi$ \citep{jaksch2010near}.}


\vspace{-2mm}
\section{The \AlgB\ Algorithm}\label{sec:algo}
\vspace{-2mm}
\subsection{Confidence Sets for Factored MDPs}
We begin with introducing empirical estimates and confidence sets used by \AlgB. Throughout this section, for each given $Z\!\subseteq\! [n]$ and $x\!\in\! \cX[Z]$, we let  $\text{N}(t, x; Z) := \max(\sum_{t'=1}^{t-1} \indic{x_{t'}[Z] = x}, 1)$ denote the number of visits to $x$ up to time $t$.

\vspace{-2mm}\noindent
\paragraph{Empirical Estimates for Factored Representation.} 
Let us consider time $t\ge 1$ and recall that $x_t = (s_t, a_t)$. 
For $i\!\in\! [m]$, we define the shorthand notation $N^p_{i,t}(x):= \text{N}(t, x; Z^p_i)$. Likewise, for $i\!\in\! [\ell]$, we define $N^r_{i,t}(x):= \text{N}(t, x; Z^r_i)$.
We then introduce the following empirical estimates of transition and reward functions. Given $i\!\in\! [m]$ and $x\!\in\! \cX[Z^p_i]$, we let $\widehat P_{i,t}(\cdot|x)$  be the empirical estimate  of $P_{i}(\cdot|x)$ built using $N^p_{i,t}(x)$ i.i.d.~samples from $P_{i}(\cdot|x)$:
\begin{align*}
\widehat P_{i,t}(y|x) &:= \frac{1}{N^p_{i,t}(x)} \sum_{t'=1}^{t-1} \indic{x_{t'}[Z^p_i] = x, s_{t'+1}[i] = y}\, .
\end{align*}
Similarly, 
given $i\in [\ell]$ and $x\in \cX[Z^r_i]$, we define $\widehat \mu_{i,t}(x)$ as the empirical estimate of $R_i(x)$ built using $N^r_{i,t}(x)$ i.i.d.~samples from $R_i(x)$: 
\begin{align*}
\widehat\mu_{i,t}(x) &:= \frac{1}{N^r_{i,t}(x)} \sum_{t'=1}^{t-1} r_{t'}[i]\indic{x_{t'}[Z^r_i] = x}\, .
\end{align*}

\vspace{-2mm}\noindent
\paragraph{Confidence Sets.}
We first define the confidence set for the reward function. For each $i\in [\ell]$ and $x\in \cX[Z_i^r]$, we introduce the following entry-wise confidence set:
\als{
c_{t,\delta,i}(x) =\bigg\{&q\in[0,1]:|\widehat \mu_{i,t}(x)  - q| \\
&\le
\sqrt{\frac{2\widehat\sigma_{i,t}^2(x)}{N^r_{i,t}(x)}\beta_{N^r_{i,t}(x)}(\delta)} \!+\! \frac{7\beta_{N^r_{i,t}(x)}(\delta)}{3N^r_{i,t}(x)}
 \bigg\}\,,
}
where $\widehat\sigma^2_{i,t}(x)$ denotes the empirical variance of the reward function $R_i(x)$ built using $N^r_{i,t}(x)$ i.i.d.~samples from $R_i(x)$, and for $n\in \NN$ and $\delta\in (0,1)$, we define 
\als{
\beta_n(\delta) &:= \eta\log\Big(\frac{\log(n)\log(\eta n)}{\log^2(\eta)\delta}\Big)\, , 
}
with $ \eta\!=\!1.12$. (In fact, any choice of $\eta>1$ is valid, however $\eta=1.12$ yields a small bound.\footnote{The optimal $\eta$ is obtained by optimizing $\beta_n(\delta)$ over $\eta$. The optimal $\eta$ will depend on $n$, but as it turns out, the optimal $\eta$ can be approximated well by a constant function $\eta=1.12$ since $\beta_n(\delta)$ grows very slowly with $n$.}) The definition of the confidence set $c_{t,\delta,i}(x)$ is obtained using an empirical Bernstein concentration inequality (see, e.g., \citet{maurer2009empirical}), modified using a peeling technique to handle arbitrary random stopping times.\footnote{We refer the interested reader to \citep{maillard2019mathematics} for the generic proof technique behind this result.} We also note that in the definition of $\beta_n(\delta)$, 
This leads us to define the following confidence set for mean rewards: For $x\in \cX$,
\als{
&\cC^r_{t,\delta}(x)\!=\!\\
&\bigg\{\!\mu'\in \cR_{\cX,[0,1]}^{\textrm{fac}}(\cG_r): \forall i\in [\ell], \mu'_i(x[Z^r_i]) \in c_{t,\delta_i,i}(x[Z^r_i]) \! \bigg\},
}
where $\delta_i = \delta(\ell |\cX[Z^r_i]|)^{-1}$.

As for the transition function, we define for each $i\!\in\! [m]$, $x\!\in\! \cX[Z^p_i]$, and $y\!\in\! \cS_i$ the following confidence set:
\als{
C_{t,\delta,i}(x,y)\!=\! \bigg\{&\! q\!\in\![0,1]\!:  | \widehat P_{i,t}(y|x) - q | \\
 &\!\le\! \sqrt{\frac{2q(1\!-\!q)}{N^p_{i,t}(x)}\beta_{N^p_{i,t}(x)}(\delta)} \!+\! \frac{\beta_{N^p_{i,t}(x)}(\delta)}{3N^p_{i,t}(x)}
\bigg\}\,.
}
This confidence set  comes from a Bernstein concentration inequality as above.\footnote{We note that \cite{bourel2020tightening} define a similar Bernstein-type confidence set for the transition function of tabular (and non-factored) MDPs.}  
Finally, we define the confidence set for $P$ as follows: For $x\in \cX$,
\als{
\cC^p_{t,\delta}(x)\!=\!\bigg\{\!P'\in \cP_{\cX,\cS}^{\textrm{fac}}&(\cG_p): \forall i\in [m], \forall y\in \cS_i,\\
&  P'_i(y|x[Z^p_i]) \in C_{t,\delta_i,i}(x[Z^p_i],y) \! \bigg\}\, ,
}
where $\delta_i = \delta(2mS_i|\cX[Z^p_i]|)^{-1}$. 
We therefore define the following set of FMDPs that are plausible at time $t$:
\als{
\cM_{t,\delta}&=\Big\{M'=(\cS,\cA,P',R') \in \mathbb M_{\cG(M)}:  \\
&\mu'(x) \in \cC^r_{t,\delta}(x) \hbox{ and } P'(\cdot|x) \in\cC^p_{t,\delta}(x), \, \forall x\in \cX\Big\}\, .
} 
By construction of the confidence sets, the set $\cM_{t,\delta}$ contains the true FMDP with high probability, and \emph{uniformly} for all time horizons $T$: Formally, $\PP\big(\exists t\!\in\! \NN, M\!\notin\! \cM_{t,\delta}\big)\leq 2\delta$. (We present a formal proof of this fact in Appendix~\ref{app:concentration_inequalities}.)

%

\vspace{-2mm}\noindent
\subsection{\AlgB: Pseudo-code}
\AlgB\ receives the structure $\cG(M)$ of the true FMDP $M$ as input. In order to implement the optimistic principle, \AlgB\ considers the set $\cM_{t,\delta}$ of plausible FMDPs and aims to compute the optimal policy $\overline\pi^+_t$ among all policies in all plausible FMDPs in $\cM_{t,\delta}$, that is $\overline{\pi}_t^+ \!=\! \argmax_{\pi:\cS\to\cA} \max\{ g_\pi^M\!:  M \!\in\! \cM_{t,\delta} \}$, where $g_\pi^M$ denotes the gain of policy $\pi$ in $M$. This maximization can be solved approximately by the Extended Value Iteration (\EVI) algorithm that builds a near-optimal policy $\pi^+_t$ and an FMDP $\widetilde M_t$ such that $g_{\pi^+_t}^{\widetilde M_t}  \geq \max_{\pi,  M\in\cM_{t,\delta} }g_\pi^M - \tfrac{1}{\sqrt{t}}$. Similarly to UCRL2 and its variants, \AlgB\ proceeds in internal episodes $k=1,2,\ldots$, where a near-optimistic policy $\pi_t^+$ is computed only at the starting time of each episode. Letting $t_k$ denote the starting time of episode $k$, the algorithm computes $\pi_k^+:=\pi_{t_k}^+$ and applies it until $t=t_{k+1}-1$, where $t_{k+1}$ is the first time step in which the number of observations gathered on some reward factor or transition factor within episode $k$ is doubled. This event writes a bit differently in the FMDP setup. Namely, the sequence $(t_k)_{k\geq 1}$ is defined as follows: $t_1=1$, and for each $k>1$
\als{
t_k  &= \min\bigg\{ t > t_{k-1}:  \max\Big\{\max_{x\in \cX[Z^p_i], i\in [m]} \frac{\nu^p_{i, t_{k-1}:t}(x)}{N^p_{i,t_{k-1}}(x)}, \\
&\max_{x\in \cX[Z^r_i], i\in [\ell]} \frac{\nu^r_{i, t_{k-1}:t}(x)}{N^r_{i,t_{k-1}}(x)}\Big\}\geq 1 \bigg\},
}
where $\nu^p_{i, t_1:t_2}(x)$ (resp.~$\nu^r_{i, t_1:t_2}(x)$) denotes the number of observations of $x\!\in\! \cX[Z^p_i]$ (resp.~of $x\!\in\! \cX[Z^r_i]$) between time $t_1$ and $t_2$. 
The pseudo-code of \AlgB\ is provided in Algorithm \ref{alg:FMDP_UCRL2B}, which uses \EVI\ (Algorithm \ref{alg:EVI}) and \texttt{InnerMax} (Algorithm \ref{alg:InMax}) as subroutines. 

\begin{algorithm}[!hbtp]
   \caption{\AlgB}
   \label{alg:FMDP_UCRL2B}
   \small
\begin{algorithmic}
\STATE \textbf{Input:} Structure $\cG$, confidence parameter $\delta$
   \STATE \textbf{Initialize:} For all $i\in [m], x\in \cX[Z^p_i]$, set $N^p_{i,0}(x)=0$. For all $i\in [\ell], x\in \cX[Z^r_i]$, set $N^r_{i,0}(x)=0$. Set $t_0=0$, $t=1$, $k=1$.
   \FOR{episodes $k=1,2,\ldots$}
       \STATE Set $t_k = t$
       \STATE Compute empirical estimates $\{\widehat \mu_{i,t_k}(x)\}_{i\in [\ell], x\in\cX[Z^r_i]}$ and $\{\widehat P_{i,t_k}(\cdot|x)\}_{i\in [m], x\in \cX[Z_i^p]}$
       \STATE Compute $\pi^+_{k} = \EVI\Big(\cM_{t_k,\delta}, \tfrac{1}{\sqrt{t_k}}\Big)$ -- see Algorithm \ref{alg:EVI} 
       \STATE Set $\nu^p_{i,k}(x)=0$ for all $i\in[m]$ and $x\in \cX[Z_i^p]$
       \STATE Set $\nu^r_{i,k}(x)=0$ for all $i\in[\ell]$ and $x\in \cX[Z_i^r]$
       \STATE \texttt{continue} = \textsf{True}
       \WHILE{\texttt{continue}}
            \STATE Observe the current state $s_{t}$, play action $a_t=\pi_{k}^+(s_t)$, and observe reward $r_t = (r_t[1], \ldots, r_t[\ell])$. Set $x_t=(s_t,a_t)$
            \STATE Set
            $
            \begin{cases}
            \nu^p_{i,k}(x_t[Z^p_i])=\nu^p_{i,k}(x_t[Z^p_i])+1, \quad i\in[m] \\
            \nu^r_{i,k}(x_t[Z^r_i])=\nu^r_{i,k}(x_t[Z^r_i])+1, \quad i\in[\ell] 
            \end{cases}
            $
            \STATE $\texttt{continue} = \bigwedge_{i\in [m]} \big(\nu^r_{i,k}(x_t[Z^r_i])< N^r_{i,t_k}(x_t[Z^r_i]\big) \wedge \bigwedge_{i\in [\ell]} \big(\nu^p_{i,k}(x_t[Z^p_i])< N^p_{i,t_k}(x_t[Z^p_i]\big) $
            \STATE Set $t=t+1$
       \ENDWHILE
        \STATE Set
        $
        \!\begin{cases}\!
        N^p_{i,t_k}(x)\!=\!N^p_{i,t_{k-1}}(x)\!+\!\nu^p_{i,k-1}(x), \, i\!\in\![m],x\!\in\!\cX[Z_i^p] \\
        N^r_{i,t_k}(x)\!=\!N^r_{i,t_{k-1}}(x)\!+\!\nu^r_{i,k-1}(x), \, i\!\in\![\ell],x\!\in\!\cX[Z_i^r] 
        \end{cases}
        $
   \ENDFOR
\end{algorithmic}
\normalsize
\end{algorithm}
%

\begin{algorithm}[!hbtp]	
	\caption{\EVI$(\cM, \epsilon)$}
	\label{alg:EVI}
    \footnotesize
\begin{algorithmic}
		\STATE Let $u_0\equiv 0, u_{-1}\equiv-\infty$, $n=0$
		\WHILE{$\max_s (u_{n}(s)-u_{n-1}(s)) -\min_s(u_{n}(s)-u_{n-1}(s)) > \epsilon$}
		\STATE $\!\!\!$Compute
		\STATE For all $(s,a)$, compute $\tilde \mu(\cdot|s,a) = \max\{\mu'(s,a):  \mu' \!\in\! \cC^r\}$.
		\STATE For all $(s,a)$, compute $\widetilde P_n(\cdot|s,a)$ using $\texttt{InnerMax}(u_n, \cC^p)$ -- See Algorithm \ref{alg:InMax}.
		\STATE $\!\!\!$Update
		$\!\begin{cases}
\!u_{n+1}\!(s) =   \max_{a\in \cA}\!\Big(\!\frac{\tilde\mu(s,a)}{\ell}\!+\!\sum_{y\in \cS}\!\widetilde P_n(y|s,a)u_n(y)\!\Big)\\
\!\pi^+_{n+1}\!(s) \!\in\! \argmax_{a\in \cA}\!\Big(\!\frac{\tilde\mu(s,a)}{\ell}\!+\!\sum_{y\in \cS}\!\widetilde P_n(y|s,a)u_n(y)\!\Big)
		\end{cases}$
		\STATE $\!\!\!n=n+1$
\ENDWHILE
	\end{algorithmic}
\normalsize
\end{algorithm}

\begin{algorithm}[!hbtp]	
	\caption{\texttt{InnerMax}$(u, \cC^p)$}
	\label{alg:InMax}
    \footnotesize
\begin{algorithmic}
		\STATE Enumerate $\cS\!=\!\{\!s_1,\!s_2,\!\ldots\!,\!s_S\!\}$ such that $u(s_1)\!\geq\!\ldots\!\geq\!u(s_{S})$
		\STATE Compute $\!\begin{cases}
P_i^+(y[i])\in \max\{q(y[i]): q\in \cC^p\}, \quad i\in[m]\\
P_i^-(y[i])\in \min\{q(y[i]): q\in \cC^p\}, \quad i\in[m]
		\end{cases}$
		\STATE Set 		
	$\!\begin{cases}
P^+(y) = \prod_{i=1}^m P^+_i(y[i])\\
		P^-(y) = \prod_{i=1}^m P^-_i(y[i])
		\end{cases}$		
		\STATE For $j\in [S]$, set $q(s_j) = \prod_{i=1}^m P_i^-(s_j[i])$
		\STATE $l = 1$
\WHILE{$\sum_{j=1}^S q(s_j) > 1$ and $l\leq S$}
			\STATE $q' = \prod_{i=1}^m P_i^{+}(s_l[i])$ 
			\STATE $q(s_l)\!=\!\max\Big(q(l)\!+\!\min\Big(1\!-\!\sum_{j=1}^S q(s_j), q'\!-\!q(s_l)\Big), 1\Big)$
			\STATE $l = l + 1$
\ENDWHILE
\STATE \textbf{Output} $q$
	\end{algorithmic}
\normalsize
\end{algorithm}

\vspace{-2mm}\noindent
\section{\AlgB: REGRET ANALYSIS}\label{sec:results}
\vspace{-2mm}\noindent
In this section, we present high-probability and finite-time regret upper bounds for \AlgB. Our main result, presented in Theorem~\ref{thm:regret_Alg1}, is a regret upper bound 
assuming a generic structure $\cG$. It is followed by Theorem \ref{thm:regret_Alg1_Cartesian} stating a substantially refined bound when the underlying structure $\cG$ admits a Cartesian product form. Before presenting the theorems, we introduce a new notion of connectivity in FMDPs.  

\vspace{-2mm}
\paragraph{Factored Diameter.}
Theorem \ref{thm:regret_Alg1} relates the regret of \AlgB\ to a new notion of connectivity in FDMPs, which we call the \emph{factored diameter}. To present its definition, for $i\in [m]$ and $x\in \cX[Z^p_i]$, we introduce: 
$\cK_{i,x}:=\supp(P_i(\cdot|x))$ and $K_{i,x} := |\cK_{i,x}|$.


\begin{definition}[Factored Diameter]
\label{def:factored_diameter}
The \emph{factored diameter} of an FMDP $M$ along factor $i\in[m]$ and for $u\in \cS[Z_i^p]$, denoted by $D_{i,u}=D_{i,u}(M)$, is defined as
\als{
D_{i,u} = \max_{s\in \cS: s[Z^p_i] = u} \max_{s_1,s_2\in \cL_{s}} \min_\pi \EE[T^\pi(s_1,s_2)]\, ,
}
where $\cL_{s} := \otimes_{i=1}^m(\cup_{a\in \cA[Z^p_i]} \cK_{i, s[Z_i^p], a})$, and where for $s_1,s_2$ with $s_1\neq s_2$, $T^\pi(s_1,s_2)$ denotes the number of steps it takes to reach $s_2$ starting from $s_1$ by following policy $\pi$. 
\end{definition}
\vspace{-2mm}

\begin{figure}[!t]
\begin{center}
	\includegraphics[scale=.7]{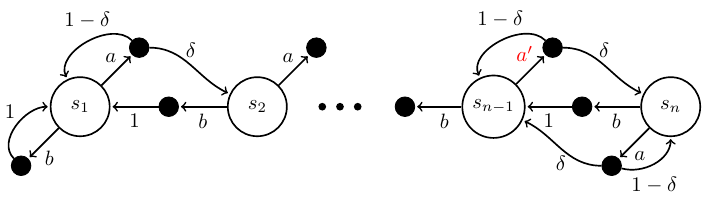}
    \caption{An Illustrative Example for the Factored Diameter.}
    \label{fig:GFdiameter}
\end{center}
\end{figure}

The notion of factored diameter refines that of diameter $D$ of \cite{jaksch2010near} (see Section \ref{sec:model}). It also extends the notion of local diameter introduced in \citep{bourel2020tightening} for tabular MDPs to FMDPs, in the sense that in the absence of factorization (i.e., when the DBN is a complete graph), the factored diameter coincides with the local diameter in \citep{bourel2020tightening}. The factored diameter takes into account the joint support sets of factors, and is therefore a problem-dependent refinement of the global diameter $D$. 
In particular, for all $i$ and $u\in \cS[Z_i^p]$, $D_{i,u} \le D$. Interestingly, there exist cases where $D_{i,y} \ll D$, as we illustrate through an example below, which is motivated by a multi-agent RL scenario. 

Consider $2$ agents, each independently interacting with an instance of the $n$-state MDP shown in Figure \ref{fig:GFdiameter}. 
Each agent $i\in\{1,2\}$ occupies a state in $\cS_i=\{s_1,s_2,\ldots,s_n\}$ with $n>2$, and the transition function $P_i$ is defined according to the MDP shown in the figure.
In each state $s\neq s_{n-1}$, each agent has access to two actions $a$ and $b$. Only when both agents are \emph{simultaneously} in $s_{n-1}$, they have access to an extra action $a'$, which causes each agent to stochastically (but independently) transit to a high-reward state $s_n$ --- For instance, this could be relevant in scenarios where cooperation yields higher rewards. 
This scenario can be modeled using an FMDP, whose state-space is $\cS=\cS_1\times \cS_2$ and whose (state-dependent) action-space is: $\cA_s=\{a,b\}\times \{a,b\}$ for all $s\neq (s_{n-1},s_{n-1})$, and $\cA_{(s_{n-1},s_{n-1})} =\{a',b\} \times \{a',b\}$.

For the case of a single agent, observe that $D=\frac{n-1}{\delta}$, as it takes $\frac{n-1}{\delta}$ steps in expectations to reach $s_n$ from $s_1$ (namely, the worst-case travel time in $\cS_i$). 
In the considered FMDP, it is easy to verify that $D=\big(\tfrac{n-1}{\delta}\big)^2$. The factored diameter here is at most $\tfrac{4}{\delta^2}$. Indeed, for $s=(s_i,s_j)$, we have $\cL_{s} = \{s_{(i-1)\vee 1},s_{i},s_{(i+1)\wedge n}\}\times\{s_{(j-1)\vee 1},s_{j},s_{(j+1)\wedge n}\}$,\footnote{We use shorthands $a\vee b\!=\!\max\{a,b\}$ and $a\wedge b\!=\!\min\{a,b\}$.} and one can verify that for any two states $u,v\in \cL_s$, it takes at most  $\tfrac{4}{\delta^2}$ steps in expectation to reach $u$ starting from $v$ -- For details see Appendix \ref{app:diameter}.

In this example, the factored diameter is smaller than $D$ by a factor of $\cO(n^2)$. Now, if we extend this to the case of $m$ agents, the ratio of $D$ and the factored diameter would be $\cO(n^m)$. This further implies that \AlgB\ achieves a much sharper regret than DORL and UCRL-Factored in these cases, whose regret bounds depend on $D$. Let us remark that such a massive reduction is a consequence of using Bernstein confidence sets for transition function that takes into account the support of $P_i$. We however stress that in contrast to non-factored MDPs where a corresponding local diameter is straightforward to define (as done in \citep{bourel2020tightening}), the task in FMDPs involves technical challenges for the decomposition of transition function along factors. To carefully exploit the gain of using Bernstein confidence intervals for $P_i$, we rely on the following factored deviation lemma, which is a refined variant of Lemma 1 in \citep{osband2014near}, and whose proof is reported in Appendix \ref{app:proof_factorization_lemma}:

\begin{lemma}
\label{lem:factored_deviation}
Let $P$ and $P'$ be two probability measures defined over $\cS=\cS_1\times\cdots\times \cS_m$ such that for all $y=(y_1,\ldots,y_m)\in \cS$: $P(y)=\prod_{i=1}^m P_i(y_i)$ and $P'(y)=\prod_{i=1}^m P'_i(y_i)$. Assume that for all $i\in [m]$, there exist $\xi_i>0$ and $\xi'_i>0$ such that 
$
|(P'_i-P_i)(y_i)|\leq \sqrt{P_i(y_i)\xi_i} + \xi'_i$, for all $y_i\in \cS_i.
$
Then, for any function $f:\cS\to \RR_+$, 
\als{
\sum_{y\in \cS} &|(P-P')(y)|f(y) \leq 3\max_{y\in \cS} f(y)\sum_{i=1}^m \xi'_iS_i \\
&+\max_{y\in\otimes_{i=1}^m\supp(P_i)} f(y) \sum_{i=1}^m \sum_{y_i\in\cS_i}\sqrt{P_i(y_i)\xi_i} \, .
}
\end{lemma}

\vspace{-2mm}
\paragraph{Regret Bound for Generic Structure.} The following theorem presents a high-probability regret bound for \AlgB\ under a generic and known structure:

\begin{theorem}[Regret of \AlgB]\label{thm:regret_Alg1}
		Uniformly over all $T\ge 3$, with probability higher than $1-\delta$, it holds that
		\als{
			&\kR(\mathrm{\AlgB},T) \leq  \cO\bigg(\!c(M)\sqrt{T \log\big(\log(T)/\delta\big)} \\
			&+\!D\!\Big(\!S_i\!\sum_{i=1}^m |\cX[Z^p_i]|\!+\!\sum_{i=1}^\ell\!|\cX[Z^r_i]|\Big)\!\log(T)\!\log\big(\log(T)/\delta\big)\!\bigg),
		}	
		with 		$c(M)=\sum\nolimits_{i\in [m]}\sqrt{\sum\nolimits_{(s,a)\in\cX[Z^p_i]} D^2_{i,s}(K_{i,s,a}-1)} +\sum\nolimits_{i\in[\ell]} \sqrt{|\cX[Z^r_i]|} + D$. 
\end{theorem}

In comparison, the regret of both UCRL-Factored and DORL satisfies $\widetilde \cO(D\sum_{i=1}^m \sqrt{S_i |\cX[Z^p_i]|T})$. 
The regret bound of \AlgB\ improves over these regret bounds as for all $i\in [m]$ and $(s,a)\in\cX[Z_i^p]$, we have $K_{i,s,a} \le S_i$ and $D_{i,s}\leq D$. In view of $D_{i,s}\ll D$ in some FMDPs, this improvement can be substantial in some domains. We also demonstrate through numerical experiments on standard environments that \AlgB\ is significantly superior to existing algorithms  that admit frequentist regret guarantees. 
We finally note that \cite{xu2020near} presented another measure called the \emph{factored span}, and present an algorithm following REGAL \citep{bartlett2009regal}, whose regret scales with the factored span (and not $D$). However, by design the presented algorithm crucially relies on knowing an upper bound on the factored span. The notions of factored diameter and factored span are not directly comparable. We however remark that the bound in Theorem \ref{thm:regret_Alg1} is achieved without any prior knowledge on the diameter. 

The proof of Theorem \ref{thm:regret_Alg1} is provided in Appendix \ref{app:regret_proof_generic}. Similarly to most UCRL2-style algorithms, the proof of this theorem follows the machinery of the regret analysis in \citep{jaksch2010near}. However, to account for the underlying factored structure, as in the regret analyses in (\citet{osband2014near,xu2020near}), the proof decomposes the regret across factors. The algorithms in  \citet{osband2014near,xu2020near} both rely on $L_1$-type confidence sets, as in UCRL2 \citep{jaksch2010near}, and their corresponding regret analyses proceed by decomposing an $L_1$ distance between two probability distributions to the sum of  $L_1$ distances over various factors. This decomposition necessarily involves the global diameter $D$ in the leading term of regret. In contrast, \AlgB\ relies on Lemma \ref{lem:factored_deviation}, which carefully exploits the benefit of using the Bernstein-style confidence sets. 
\vspace{-2mm}

\paragraph{Regret Bound for Cartesian Products.} 
We now focus on a structure $\cG$ that can be represented as a Cartesian product, so that the true FMDP $M$ can be seen as a Cartesian product of some base MDPs. Let the true FMDP $M$ be a Cartesian product of $m$ base MDPs, $M_i, i=1,\ldots,m$, each with state-space $\cS_i$, action-space $\cA_{i}$, and diameter $D_i$. 

\vspace{-2mm}
\begin{theorem}[Regret of \AlgB\ for Cartesian products]\label{thm:regret_Alg1_Cartesian}

		With probability higher than $1-\delta$, for all $T\ge 3$,
		\als{
			\kR&(\mathrm{\AlgB},T) \leq  \cO\bigg( \sum_{i=1}^m c_i\sqrt{T \log\big(\log(T)/\delta\big)} \\
			& + \sum_{i=1}^m D_iS_iA_i\log(T)\log\big(\log(T)/\delta\big)\bigg)\,, \,\, \text{with}
		}	
		$c_i\!=\!\sqrt{\sum\nolimits_{i=1}^m\sum\nolimits_{s\in \cS_i, a\in \cA_i} D_{i,s}^2K_{i,s,a}}\!+\!\sum\nolimits_{i=1}^m \sqrt{S_iA_i} + D_{i}$. 
\end{theorem}

\vspace{-2mm}
This result asserts that in the case of Cartesian products, the regret of \AlgB\ boils down to the sum of individual regret of $m$ base MDPs, where each individual term corresponds to a fully local quantity, i.e.~depending only on the properties of $M_i$. This bound significantly improves over previous regret bounds for the product case, which were unable to establish a fully localized regret bound. In particular, the bounds of \citep{osband2014near} and \citep{xu2020near} for this case would necessarily depend on the {\it global diameter of the FMDP}, which might scale as $\prod_{i=1}^m D_i$, whereas ours in Theorem \ref{thm:regret_Alg1_Cartesian} depends on {\it the local diameter of the local MDPs}. This would in turn imply an exponentially (in the number $m$ of base MDPs) tighter regret bound. 
Finally, we mention that \cite{xu2020near} present a regret lower bound scaling as $\Omega(\sqrt{bLT})$ in FMDPs based on worst-case Cartesian products, where $L$ is an upper bound on both $|\cX[Z_i^p]|$ and $|\cX[Z_i^r]|$, and $\mathrm{b}$ denotes the span of the optimal bias function. Our regret bounds do not contradict this lower bound as $b\le \sum_{i} D_{i,s}$ for any $s$.  

\vspace{-2mm}
\begin{remark}
Cartesian products might seem specific, but admittedly they represent the extreme case of FMDPs, where the individual MDPs are independent of one another. Hence, they are used in existing works (e.g., \cite{osband2014near,xu2020near}) to establish \emph{best-case bounds} on exploration. The resulting bounds are typically more explicit than their corresponding complicated bounds for generic FMDPs. This allowed us to establish a best-case bound depending only in \emph{fully local} quantities, in contrast to existing bounds above for Cartesian products. We believe that there is still value in analysing these special cases, and that analysing intermediate cases (in which individual MDPs are only \emph{weakly connected}) is an important avenue for future work. 
\end{remark}

\vspace{-2mm}
We finally note that Theorem \ref{thm:regret_Alg1_Cartesian} cannot be directly obtained from Theorem \ref{thm:regret_Alg1}, and its proof crucially relies on the following lemma stating that in FMDPs with Cartesian structures, the value function can be decomposed into the sum of individual value functions of the base MDPs:\footnote{A similar results for the bias function of the FMDP in the case of Cartesian product was provided in  \cite{xu2020near}, but not for Value Iteration (VI).} 

\vspace{-2mm}
\begin{lemma}[VI for Cartesian products] \label{lem:VI_Cartesian_prod}
Consider VI with $u_0(s) = 0$, and for each $n\ge 0$, $u_{n+1}(s)=\max_{a\in \cA}\big\{m^{-1}\mu(s,a) + \sum_{y\in \cS} P(y|s,a) u_n(y)\big\}$. Then, for all $n$, $u_{n}(s) = m^{-1}\sum_{i=1}^m u^{(i)}_{n}(s[i])$, where $(u^{(i)}_{n})_{n\geq 0}$ is a sequence of VI on MDP $i$, that is $u^{(i)}_0(x)=0$ and $u^{(i)}_{n+1}(x) = \max_{a\in \cA_i}\big\{\mu_i(x,a) +  \sum_{y\in \cS_i} P_i(y|x,a) u^{(i)}_{n}(y)\big\}$ for all $x\in \cS_i$ and $n\geq 0$. 
\end{lemma}


\vspace{-2mm}
\section{\uppercase{Numerical Experiments}}\label{sec:xps}
\vspace{-2mm}
In this section, we present results from numerical experiments with \AlgB. We perform experiments with the algorithm in two domains: Two factored versions of \emph{RiverSwim} \citep{strehl2008analysis,filippi2010optimism}, and the \emph{SysAdmin} domain \citep{guestrin2003fmdps}.  

We consider two factored versions of RiverSwim. In the first one, we construct an FMDP by taking the Cartesian product of two RiverSwim instances, with $6$ states each, and introduce additional reward for a single joint state to couple the two instances through the reward factors. This corresponds to $S=36$ and $A=4$ (and so, $|\cX|=144$). We shall refer to this domain as Two-Layer RiverSwim. We construct the second factored version of RiverSwim by 
coupling three RiverSwim instances with $4$ states each, in a similar fashion. This results in an FMDP with $S=64$ and $A=8$ (and hence, $|\cX|=512$), which we call Three-Layer RiverSwim. 
Recall that $\cX = \cS_1\otimes\cdots\otimes\cS_m\otimes\cA_1\otimes\cdots\otimes\cA_{n-m}$, i.e.~each factor scope $\cX[Z_i^p]$ (resp.~$\cX[Z_i^r]$) is the Cartesian product of a subset of state and action factors. In other words, the agent knows which subset of factors is relevant for transition factor $P_i$ (resp.~reward factor $R_i$), but does not have access to a compact representation, e.g.~in the form of a decision tree. Having access to such a compact representation would improve the performance of the algorithm but makes a stronger assumption on the available prior knowledge.

We compare \AlgB{} to the following three algorithms\footnote{The code is made publicly available via 
\url{https://github.com/aig-upf/dbn-ucrl}.}: UCRL-Factored \citep{osband2014near}, the previous state-of-the-art algorithm for FMDPs, which is the natural extension of UCRL2 \citep{jaksch2010near} to FMDPs; PSRL-Factored \citep{osband2014near}, which is an algorithm based on posterior sampling and is in fact a natural extension of PSRL \citep{osband2013more} to episodic FMDPs; and 
UCRLB-peeling, an improved  variant of UCRL2 and UCRL2B \citep{fruit2020improved} relying on the same confidence sets for reward and transition functions as in \AlgB{} but ignoring the factored structure. 
In particular, the comparison against UCRLB-peeling indicates the gain achieved by taking into account the factored structure, whereas that against UCRL-Factored reveals the gain of (element-wise) Bernstein-type confidence sets over their counterparts derived using Hoeffding's and Weissman's concentrations. 
As far as we know, ours is the first full-scale empirical evaluation of regret minimization algorithms for FMDPs. 
We stress that among these algorithms, PSRL-Factored is only shown to guarantee a Bayesian regret bound, and to the best of our knowledge, its frequentist regret analysis is still open. (We also refer to \citep{xu2020near} for a Bayesian regret analysis of PSRL-Factored in the average-reward setting.) Finally, to ease its implementation, in our experiments we let PSRL-Factored have access to the reward function. 


%
%
%
%
%

\begin{figure}
	\centering
	\includegraphics[scale=.18]{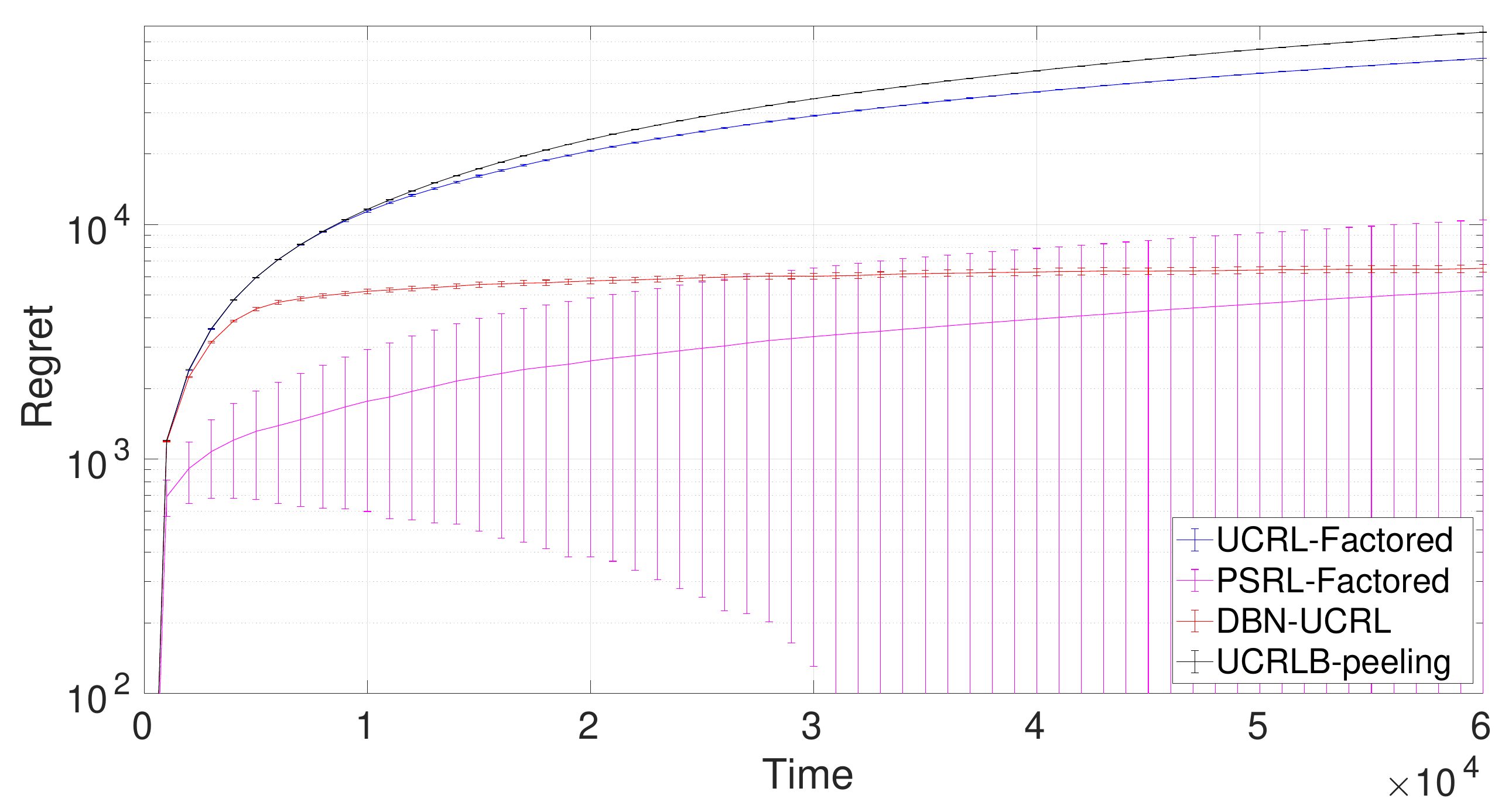}
	\caption{Regret in Two-Layer RiverSwim.}
	\label{fig:TwoRiverSwim}
\end{figure}

In our experiments, we set $\delta=0.01$ and report for each domain the average results over 50-100 independent experiments (depending on the domain), along with 95\% confidence intervals. 
Figure~\ref{fig:TwoRiverSwim} shows the regret of various algorithms against time in Two-Layer RiverSwim. (Note the logarithmic scale on the y-axis.) As the figure reveals, the regret under \AlgB\ significantly improves over that of UCRL-Factored and UCRLB-peeling, and remains competitive with PSRL-Factored. However, we observe that the regret under PSRL-Factored has a very large variance, which is in stark contrast to the other algorithms. Finally note that both \AlgB\ and PSRL-Factored enjoy a short \emph{burn-in} phase (compared to UCRLB-peeling and UCRL-Factored), after which the regret grows sublinearly with time.

\begin{figure}
	\centering
	\includegraphics[scale=.18]{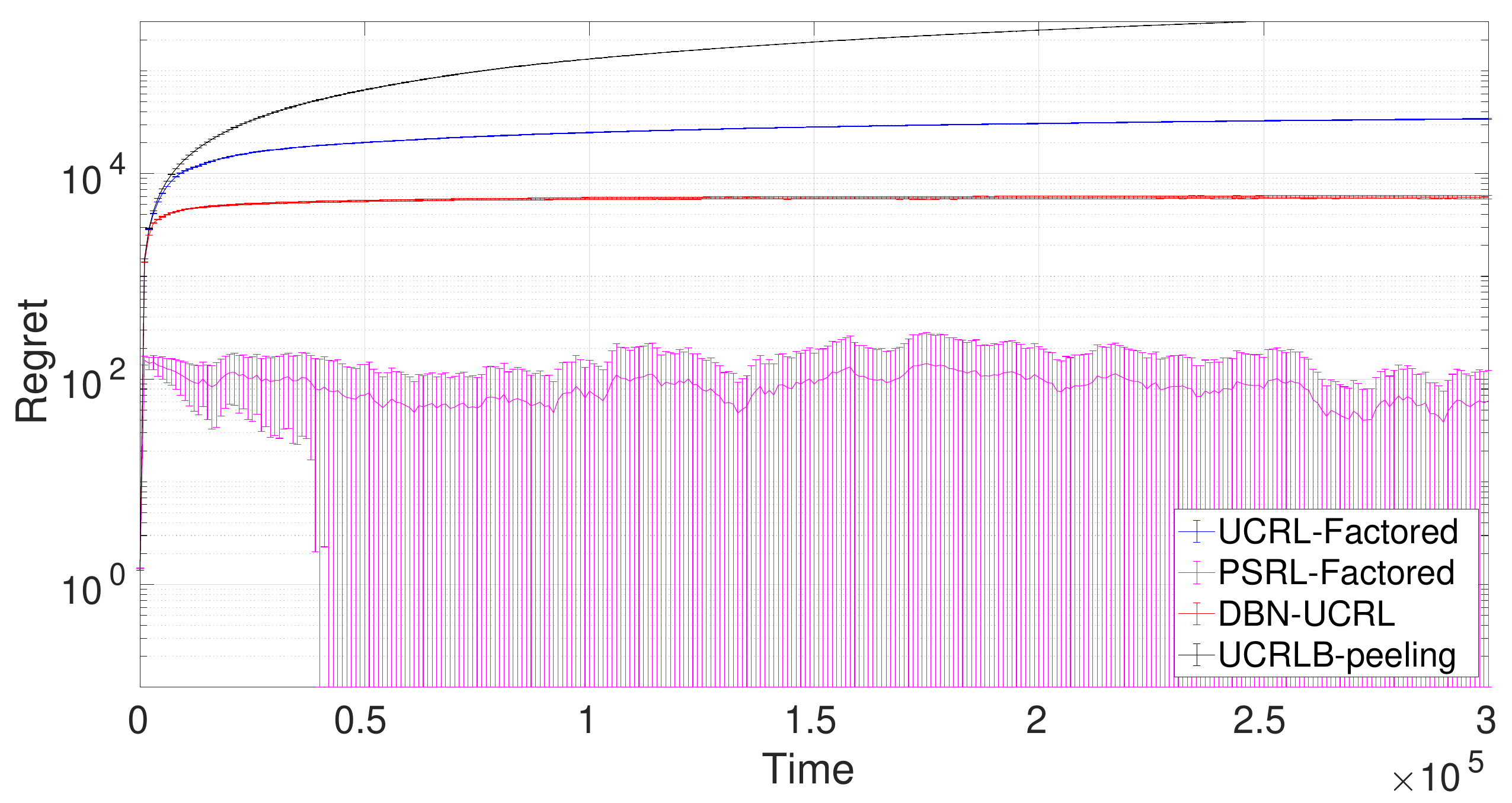}
	\caption{Regret in Three-Layer RiverSwim.}
	\label{fig:ThreeRiverSwim}
\end{figure}

Figure~\ref{fig:ThreeRiverSwim} displays the regret of various algorithms against time in Three-Layer RiverSwim. The regret under \AlgB\ significantly improves over that of UCRL-Factored and UCRLB-peeling, but is considerably worse than that of PSRL-Factored. Again, we see that the regret under PSRL-Factored has a high variance, although its average regret is smaller than the rest.



We perform two experiments in the SysAdmin domain. This domain consists of $N$ computer servers that are organized in a graph with a certain topology. Each server is represented by a binary variable that indicates whether or not it is working. At each time step, each server has a chance of failing, which depends on its own status and the status of the servers connected to it. There are $N+1$ actions: $N$ actions for rebooting a server (after which it works with high probability) and an idle action. In previous work, researchers have performed experiments with two different topologies: A circular topology in which each server is connected to the next server in the circle, and a three-legged topology in which the servers are organized in a tree with three branches. In each topology, the status of each server depends on at most one other server.

\begin{figure}
	\centering
	\includegraphics[scale=.18]{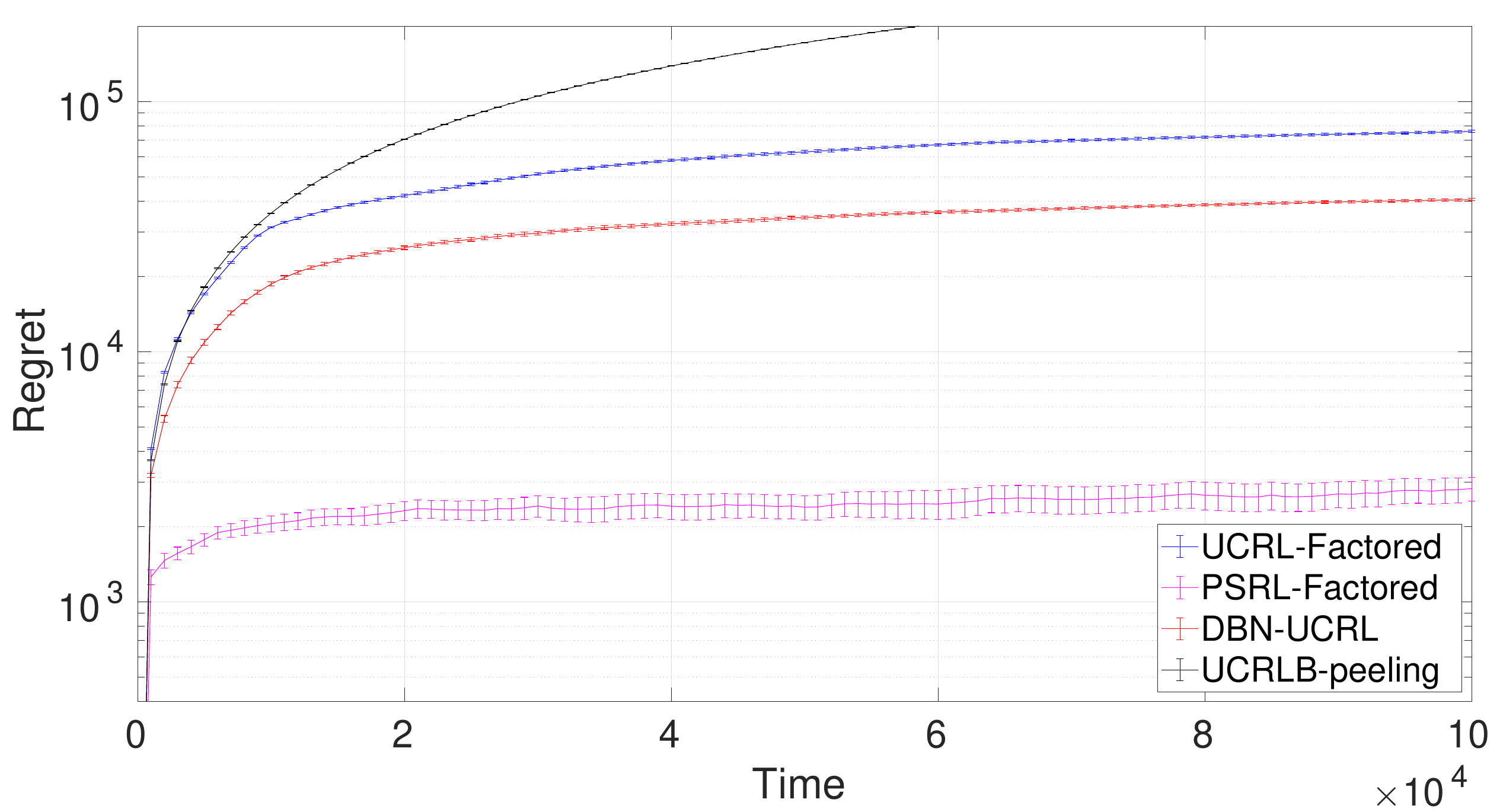}
	\caption{Regret in SysAdmin with the circular topology.}
	\label{fig:circle-regret}
\end{figure}

\begin{figure}
	\centering
	\includegraphics[scale=.18]{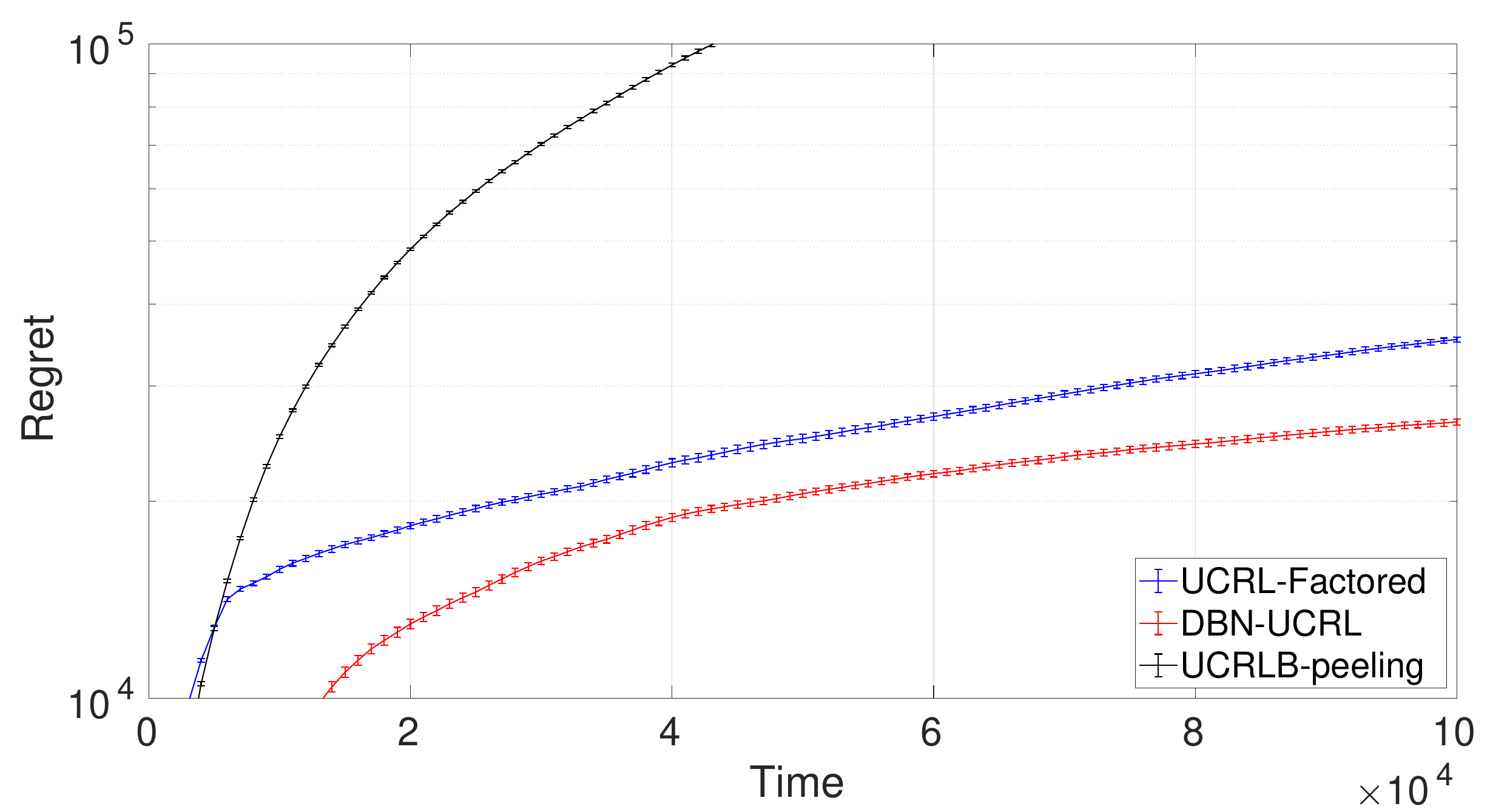}
	\caption{Regret in SysAdmin with the three-legged topology.}
	\label{fig:threeleg-regret}
\end{figure}

Figures~\ref{fig:circle-regret} and \ref{fig:threeleg-regret} show the regret of various algorithms in the SysAdmin domain for the two topologies, along with 95\% confidence intervals. For each topology, $N=7$, i.e.~the circular topology has 7 servers arranged in a circle, and the three-legged topology has a root server and two servers on each of the three branches. Hence, the respective size of the state and action-space is $S=2^7=128$ and $A=8$, and so, $|\cX|=1024$. Note again the logarithmic scale on the y-axis. As in the other domains, \AlgB{} clearly outperforms the other algorithms in terms of regret, but does worse than PSRL-Factored. (PSRL-Factored has a similar performance in both SysAdmin domains, so we only reported its regret for one.)  Compared to the previous RiverSwim domains, the regret under PSRL-Factored has a much smaller variance.    

In summary, in these experiments, \AlgB\ significantly outperformed existing algorithms for which high-probability frequentist regret bounds exist. Furthermore, it incurred a worse average regret than  PSRL-Factored in most domains, but the latter was shown to suffer from a large variance -- In contrast to confident behavior of \AlgB. We again remark that PSRL-Factored is only shown, to our knowledge, to guarantee a Bayesian regret bound, which is weaker than the corresponding high-probability frequentist regret bound. 

\vspace{-2mm}
\section{\uppercase{Conclusions}}
\vspace{-2mm}
We studied reinforcement learning under the average-reward criterion in a Factored Markov Decision Process (FMDP) with a known factorization structure, and introduced \AlgB, an optimistic algorithm maintaining Bernstein-type confidence sets for individual elements of transition probabilities for each factor. We presented two high-probability regret bounds for \AlgB, strictly improving existing regret bounds: The first one is valid for any factorization structure making appear the notion of factored diameter for FMDPs, whereas the second concerns structures taking the form of a Cartesian product. We also demonstrated through numerical experiments on standard
environments that \AlgB\ enjoys a significantly superior empirical regret than existing algorithms that admit frequentist regret guarantees. 
One interesting future direction is to derive regret lower bounds valid for FMDPs with a generic structure. 

\subsubsection*{Acknowledgements}
The authors would like to thank anonymous reviewers for their comments. Anders Jonsson is partially supported by Spanish grants PID2019-108141GB-I00 and PCIN-2017-082. Odalric-Ambrym Maillard  is supported by 
CPER Nord-Pas-de-Calais/FEDER DATA Advanced data science and technologies 2015-2020, the French Ministry of Higher Education and Research, Inria, Inria Scool, the French Agence Nationale de la Recherche (ANR) under grant ANR-16-CE40-0002 (the BADASS project), the MEL, the I-Site ULNE regarding  project R-PILOTE-19-004-APPRENF.


\bibliography{Bandit_RL_bib}
\bibliographystyle{plainnat}


\appendix
\onecolumn


\section{\uppercase{Proof of Factorization Lemma (Lemma}  \ref{lem:factored_deviation})}
\label{app:proof_factorization_lemma}

In this section, we prove Lemma \ref{lem:factored_deviation} (restated below), which provides a bound on factored deviations, and can be seen as a refined variant of Lemma 1 in \cite{osband2014near}. 


\textbf{Lemma 1 (Restated)} 
\emph{
Let $P$ and $P'$ be two probability measures defined over $\cS=\cS_1\times\cdots\times \cS_m$ such that for all $y=(y_1,\ldots,y_m)\in \cS$: $P(y)=\prod_{i=1}^m P_i(y_i)$ and $P'(y)=\prod_{i=1}^m P'_i(y_i)$. Assume that for all $i\in [m]$, there exist positive numbers $\xi_i$ and $\xi'_i$ such that 
$$
|(P'_i-P_i)(y_i)|\leq \sqrt{P_i(y_i)\xi_i} + \xi'_i\, , \quad \forall y_i\in \cS_i.
$$
Then, for any function $f:\cS\to \RR_+$, we have
\als{
\sum_{y\in \cS} |(P-P')(y)|f(y) &\leq \max_{y\in\otimes_{i=1}^m\supp(P_i)} f(y) \sum_{i=1}^m \sum_{y_i\in\cS_i}\sqrt{P_i(y_i)\xi_i} + 3\max_{y\in \cS} f(y)\sum_{i=1}^m \xi'_iS_i\, ,
}
where for a distribution $q$, $\supp(q)$ denotes the support set of $q$. 
}

\bp
We prove the lemma by induction on $m$. 
For $m=2$, we have
\als{
\sum_{y\in \cS} |(P-P')(y)|f(y) &\leq \sum_{y_1}\sum_{y_2} |P_1(y_1)P_2(y_2) - P'_1(y_1)P'_2(y_2)|f(y) \\
&\leq \sum_{y_1}\sum_{y_2} P_1(y_1)|P_2(y_2) - P'_2(y_2)| f(y)+ \sum_{y_1}\sum_{y_2} P'_2(y_2)|P_1(y_1) - P'_1(y_1)| f(y) \, .
}
The first term is bounded as:
\als{
\sum_{y_1}\sum_{y_2} P_1(y_1)|P_2(y_2) &- P'_2(y_2)| f(y) \leq  \sum_{y_2} |P_2(y_2) - P'_2(y_2)| \max_{y_1\in \supp(P_1)} f(y)\underbrace{\sum_{y_1} P_1(y_1)}_{=1} \\
&\leq  \sum_{y_2} \max_{y_1\in \supp(P_1)} f(y)\Big(\sqrt{P_2(y_2)\xi_2} + \xi'_2\Big) \\
&\leq \max_{y_1\in \supp(P_1),y_2\in\supp(P_2)} f(y) \sum_{y_2} \sqrt{P_2(y_2)\xi_2} + \xi'_2 S_2\max_y f(y) \, .
}
For the second term, we have:
\als{
\sum_{y_1}\sum_{y_2} &P'_2(y_2)|P_1(y_1) - P'_1(y_1)| f(y)  \\
&= \sum_{y_1}|P_1(y_1) - P'_1(y_1)| \bigg(\sum_{y_2\in \mathrm{supp}(P_2)} P'_2(y_2)f(y) 
+ \sum_{y_2\notin \mathrm{supp}(P_2)} P'_2(y_2)f(y) \bigg)\\
&\leq \sum_{y_1}|P_1(y_1) - P'_1(y_1)| \Big(\max_{y_2\in\supp(P_2)} f(y) \sum_{y_2\in \mathrm{supp}(P_2)} P'_2(y_2) 
+ \max_{y}f(y) \sum_{y_2\notin \mathrm{supp}(P_2)} P'_2(y_2)\Big)\\
&\stackrel{(\mathrm{a})}\leq \sum_{y_1}|P_1(y_1) - P'_1(y_1)| \Big(\max_{y_2\in\supp(P_2)} f(y)
+ \xi'_2S_2\max_{y}f(y)\Big)\\
&\stackrel{(\mathrm{b})}\leq \sum_{y_1} \Big(\sqrt{P_1(y_1)\xi_1} + \xi'_1\Big) \max_{y_2\in\supp(P_2)} f(y) + 2\xi'_2S_2\max_{y}f(y)  \\
&\leq \max_{y_1\in \supp(P_1),y_2\in\supp(P_2)} f(y) \sum_{y_1} \sqrt{P_1(y_1)\xi_1} + (\xi'_1S_1 + 2\xi'_2S_2)\max_{y}f(y)  \,,
}
where (a) follows from the fact that $P'_2(x)\leq \xi'_2$ for all $x\notin \mathrm{supp}(P_2)$, and where (b) uses $\sum_{y_1} |(P_1 - P'_1)(y_1)| \leq 2$. Hence,
\als{
\sum_{y\in \cS} |(P-P')(y)|f(y) &\leq \max_{y_i\in \otimes_{i=1}^2\supp(P_i)} f(y)  \sum_{i=1}^2\sum_{y_i} \sqrt{P_i(y_i)\xi_i} + 3\max_{y}f(y) \sum_{i=1}^2 \xi'_iS_i\, .
}

Now assume that the induction hypothesis is correct for $m>2$:
\als{
\sum_{y\in \cS} |(P-P')(y)|f(y)  &\leq \max_{y_i\in \otimes_{i=1}^m\supp(P_i)} f(y)  \sum_{i=1}^m\sum_{y_i} \sqrt{P_i(y_i)\xi_i} + 3\max_{y}f(y) \sum_{i=1}^m \xi'_iS_i\, .
}
We then show that the above holds for $m+1$. To this aim, we define shorthand $y_{1:m} := y_1,\ldots,y_{m}$, and let $q(y_{1:m})=P_1(y_1)\cdots P_m(y_m)$ and $q'(y_{1:m})=P'_1(y_1)\cdots P'_m(y_m)$. We have:
\als{
\sum_{y} |(P-P')|&(y)f(y) = \sum_{y} \Big|P_{m+1}(y_{m+1})q(y_{1:m}) - P'_{m+1}(y_{m+1})q'(y_{1:m})\Big| f(y) \\ 
&\leq \sum_{y_{1:m}}\sum_{y_{m+1}} q(y_{1:m})|(P_{m+1} - P'_{m+1})(y_{m+1})| f(y)  +\sum_{y_{1:m}}\sum_{y_{m+1}} P'_{m+1}(y_{m+1})|(q - q')(y_{1:m})| f(y) \, .
}
The first term is bounded as:
\al{
\sum_{y_{m+1}}\sum_{y_{1:m}} q(y_{1:m})|(P_{m+1} &- P'_{m+1})(y_{m+1})| f(y)  \leq  \sum_{y_{m+1}} |(P_{m+1} - P'_{m+1})(y_{m+1})| \max_{y_{1:m}\in \otimes_{i=1}^{m} \mathrm{supp}(P_i)} f(y)  \underbrace{\sum_{y_{1:m}} q(y_{1:m})}_{=1} \sk
&\leq  \sum_{y_{m+1}} \Big(\sqrt{P_{m+1}(y_{m+1})\xi_{m+1}} + \xi'_{m+1}\Big) \max_{y_{1:m}\in \otimes_{i=1}^m \mathrm{supp}(P_i)} f(y)  \sk
\label{eq:1st_term}
&\leq  \max_{y\in \otimes_{i=1}^{m+1} \mathrm{supp}(P_i)} f(y)  \sum_{y_{m+1}} \sqrt{P_{m+1}(y_{m+1})\xi_{m+1}} + \xi'_{m+1}S_{m+1} \max_y f(y) 
 \, .
}
The second term is bounded as follows:
\al{
&\sum_{y_{m+1}}\sum_{y_{1:m}} P'_{m+1}(y_{m+1})|(q - q')(y_{1:m})| f(y)   \sk
&\leq \sum_{y_{1:m}} |(q - q')(y_{1:m})|  \Big(\sum_{y_{m+1}\in \mathrm{supp}(P_{m+1})} P'_{m+1}(y_{m+1})f(y)  
+ \sum_{y_{m+1}\notin \mathrm{supp}(P_{m+1})} P'_{m+1}(y_{m+1})f(y) \Big)\sk
\label{eq:2nd_term}
&\leq \sum_{y_{1:m}}|(q - q')(y_{1:m})| \max_{y_{m+1}\in \mathrm{supp}(P_{m+1})} f(y) 
+ 2\xi'_{m+1}S_{m+1}\max_{y} f(y)  \,.
}
Note that the induction hypothesis implies 
\als{
\sum_{y_{1:m}}|(q - q')(y_{1:m})| \max_{y_{m+1}\in \mathrm{supp}(P_{m+1})} f(y)  &\leq \max_{y_i\in \otimes_{i=1}^{m+1}\supp(P_i)} f(y)  \sum_{i=1}^{m}\sum_{y_i} \sqrt{P_i(y_i)\xi_i} 
+ 3\max_{y}f(y) \sum_{i=1}^m \xi'_iS_i\, .
}
Putting this together with (\ref{eq:2nd_term}), and combining with (\ref{eq:1st_term}) yield the desired result:
\als{
\sum_{y} |(P-P')(y)|f(y) &\leq \max_{y_i\in \otimes_{i=1}^{m+1}\supp(P_i)} f(y)  \sum_{i=1}^{m+1}\sum_{y_i} \sqrt{P_i(y_i)\xi_i} + 3\max_{y}f(y) \sum_{i=1}^{m+1} \xi'_iS_i\, ,
}
thus concluding the proof. 
\ep

\section{\uppercase{Concentration Inequalities}}
\label{app:concentration_inequalities}
In this section, for the sake of completeness, we collect some concentration inequalities used when constructing the set $\cM_{t,\delta}$ of plausible MDPs in \AlgB. The first lemma provides a time-uniform Bernstein-type concentration inequality for bounded random variables: 

\begin{lemma}
\label{lem:Bernstein_peeling}
  Let $Z = (Z_t)_{t\in \NN}$ be a sequence of random  variables generated by a predictable process, and $\cF=(\cF_t)_{t}$ be its natural filtration. Assume for all $t\in \NN$, $|Z_t|\le b$ and $\EE[Z_{s}^2|\cF_{s-1}] \le v$ for some positive numbers $v$ and $b$.
  Let $n$ be an integer-valued (and possibly unbounded) random variable that is $\cF$-measurable. Then, for all $\delta\in (0,1)$,
  \als{
\PP\bigg[\exists n\in \NN, \,\, \frac{1}{n}\sum_{t=1}^{n}Z_t
\geq \sqrt{\frac{2\beta_n(\delta) v}{n}} +  \frac{\beta_n(\delta)b}{3n} \bigg] &\leq \delta\, , \\
\PP\bigg[\exists n\in \NN, \,\, \frac{1}{n}\sum_{t=1}^{n}Z_t \leq -\sqrt{\frac{2\beta_n(\delta) v}{n}} -  \frac{\beta_n(\delta)b}{3n} \bigg]  &\leq \delta\,  ,
}
where $\beta_{n}(\delta):= \eta\log\Big(\frac{\log(n)\log(\eta n)}{\delta \log^2(\eta)}\Big)$, with $\eta>1$ being an arbitrary parameter.
\end{lemma}

The next lemma presents a time-uniform concentration for i.i.d.~random variables supported in $[0,1]$:

\begin{lemma}
\label{lem:bounded_peeling}
 Let $Z = (Z_t)_{t\in \NN}$ be a sequence of i.i.d.~random  variables bounded in $[0,1]$, with mean $\mu$. Then, for all $\delta\in(0,1)$, it holds 
\als{
&\PP\bigg[\exists n\in \NN, \,\, \mu - \frac{1}{n}\sum_{t=1}^{n}Z_t \geq \sqrt{\frac{\beta_n(\delta)}{2n}} \bigg] \le \delta \, ,\\  
&\PP\bigg[\exists n\in \NN, \,\, \mu - \frac{1}{n}\sum_{t=1}^{n}Z_t \leq -\sqrt{\frac{\beta_n(\delta)}{2n}} \bigg] \le \delta\, ,
}
where $\beta_{n}(\delta):= \eta\log\Big(\frac{\log(n)\log(\eta n)}{\delta \log^2(\eta)}\Big)$, with $\eta>1$ being an arbitrary parameter.
\end{lemma}

Both Lemma \ref{lem:Bernstein_peeling} and Lemma \ref{lem:bounded_peeling} are derived from Lemma~2.4 in \citep{maillard2019mathematics}.

We also present the following lemma implying that the set of MDPs $\cM_{t,\delta}$ contains the true MDP by high probability:

\begin{lemma}\label{lem:CI_has_trueMDP}
For any FMDP with rewards in $\cR^{\text{fac}}_{\cX,[0,1]}(\cG_r)$, and transition function in $\cP^{\text{fac}}_{\cX,\cS}(\cG_p)$, for all $\delta\in(0,1)$, it holds
\als{
&\Pr\bigg(\exists t\in\Nat, x\in \cX, \quad  \mu(x) \notin \cC^r_{t,\delta}(x)
	\,\,\text{ or } \,\,P(\cdot|x) \notin \cC^p_{t,\delta}(x)\bigg)\leq 2\delta\,.
}	
In particular, for all $T\in \NN$: $\PP\big(\exists t\in \NN: M\notin \cM_{t,\delta}\big) \le 2\delta$. 	
\end{lemma}

\bp
First note that for any $i\in [\ell]$ and $x\in \cX[Z_i^r]$, by a time-uniform version of \citep[Theorem~10]{maurer2009empirical}:
$
\PP\big(\exists t\in \NN: \mu_i(x) \notin c_{t,\delta,i}(x)\big) \leq \delta. 
$
Using union bounds and recalling that $\delta_i=\delta(\ell|\cX[Z_i^r]|)^{-1}$, it then follows that
\als{
\Pr\big(\exists t\in\Nat, x\in \cX,  \mu(x) \notin \cC^r_{t,\delta}(x)\big)  &\leq \Pr\big(\exists t\in\Nat, i\in [\ell], x\in \cX[Z_i^r], \mu_i(x) \notin c_{t,\delta_i,i}(x)\big) \\
&\leq \sum_{i\in [\ell]}\sum_{x\in \cX[Z_i^r]} \PP\big(\exists t\in \NN: \mu_i(x) \notin c_{t,\delta_i,i}(x)\big) \\
&\leq \sum_{i\in [\ell]}\sum_{x\in \cX[Z_i^r]}  \frac{\delta}{\ell|\cX[Z_i^r]|} = \delta.
}	

Now we consider the case of transition function. 
For any $i\in [m]$, $x\in \cX[Z_i^m]$, and $y\in \cS_i$, Lemma \ref{lem:Bernstein_peeling} implies:
$$
\PP\big(\exists t\in \NN: P_i(y|x) \notin C_{t,\delta,i}(x,y)\big) \leq 2\delta\, . 
$$
Using union bounds gives
\als{
\Pr\big(\exists t\in\Nat, x\in \cX, \,  P(\cdot|x) \notin \cC^p_{t,\delta}(x)\big)  &\leq \Pr\big(\exists t\in\Nat, i\in [m], x\in \cX[Z_i^p], y\in \cS_i, \, P_i(y|x) \notin C_{t,\delta_i,i}(x,y)\big) \\
&\leq \sum_{i\in [m]}\sum_{x\in \cX[Z_i^p]}\sum_{y\in \cS_i} \PP\big(\exists t\in \NN: P_i(y|x) \notin C_{t,\delta_i,i}(x,y)\big) \\
&\leq \sum_{i\in [m]}\sum_{x\in \cX[Z_i^p]}\sum_{y\in \cS_i} \frac{2\delta}{2mS_i|\cX[Z^p_i]|} = \delta.
}	
Putting together and taking a union bound complete the proof. 
\ep

\section{\uppercase{Regret Analysis: Generic Structures (Theorem} \ref{thm:regret_Alg1})}
\label{app:regret_proof_generic}

In this section, we prove Theorem \ref{thm:regret_Alg1}. Our proof follows similar lines as in the proof of \citep[Theorem~2]{jaksch2010near}. Let $\delta\in (0,1)$. To simplify notations, let us define the shorthand $J_k:=J_{t_k}$ for a generic measurable random variable $J$ that is fixed within a given episode $k$ and omit its dependence on $\delta$ (for example, $\cM_{k}:=\cM_{t_k,\delta}$). Denote by $K(T)$ the number of episodes initiated by the algorithm up to time $T$. {\color{black}Given a pair $x=(s,a)$, let $N_t(x)$ denote the number of times $x$ is visited by the algorithm up to time $t$. Furthermore, let $n_k(x)$ denote the number of times $x$ is sampled in a given episode $k$.}


By applying Lemma \ref{lem:bounded_peeling} and noting that $r_t[i]\in [0,1]$, we deduce that
\als{
\kR(T)&= \sum_{t=1}^T g^\star- \sum_{t=1}^T \frac{1}{\ell}\sum_{i=1}^\ell r_t[i] \leq \sum_{x\in \cX} N_{T}(x)\Big(g^\star -  \frac{1}{\ell}\sum_{i=1}^\ell \mu_i(x[Z_i^r])\Big) + \sqrt{T\beta_T(\delta)/2}\, ,
}
with probability at least $1-\delta$. We have
\als{
\sum_{x} N_{T}(x)\Big(g^\star - \frac{1}{\ell}\sum_{i=1}^\ell \mu_i(x)\Big) &= \sum_{k=1}^{K(T)}  \sum_{x}  \sum_{t=t_k}^{t_{k+1}-1}\indic{x_t=x} \Big(g^\star - \frac{1}{\ell}\sum_{i=1}^\ell \mu_i(x[Z_i^r])\Big)\\
&=
\sum_{k=1}^{K(T)} \sum_{x} n_k(x) \Big(g^\star - \frac{1}{\ell}\sum_{i=1}^\ell \mu_i(x[Z_i^r])\Big)\,.
}
Introducing $\Delta_k := \sum_{x\in \cX} n_k(x) \big(g^\star - \frac{1}{\ell}\sum_{i=1}^\ell \mu_i(x[Z_i^r])\big)$ for $1\leq k\leq K(T)$, we get
\als{
\kR(T) \le \sum_{k=1}^{K(T)} \Delta_k + \sqrt{T\beta_T(\delta)/2} \,, 
}
with probability at least $1-\delta$. A given episode $k$ is called \emph{good} if $M \in \cM_{k}$ (that is, the set of plausible MDPs contains the true model), and \emph{bad} otherwise. By Lemma \ref{lem:CI_has_trueMDP}, for all $T$, and for all episodes $k=1,\ldots,K(T)$, the set $\cM_{k}$ contains the true MDP with probability higher than $1-2\delta$. As a consequence, with probability at least $1-2\delta$, $
\sum_{k=1}^{K(T)}\Delta_k\indic{M \notin \cM_{k}} = 0.
$

To upper bound regret in good episodes, we closely follow \citep{jaksch2010near} and decompose the regret to control the deviation of optimistic transition and reward functions from their true values. Consider a good episode $k$ (hence, $M \in \cM_{k}$). By choosing $\pi^+_{k}$ and $\widetilde M_{k}$, we get that
$$
g_k := g_{\pi^+_{k}}^{\widetilde M_{k}} \geq g^\star - \frac{1}{\sqrt{t_k}}\, ,
$$
so that with probability greater than $1-\delta$,
\begin{align}
\Delta_k  &\leq \sum_{x\in \cX} n_k(x)
\Big(g_k - \frac{1}{\ell}\sum_{i=1}^\ell \mu_i(x[Z_i^r])\Big) + \sum_{x\in \cX} \frac{n_k(x)}{\sqrt{t_k}} 
\, .
\label{eq:delta_init}
\end{align}
Using the same argument as in the proof of \citep[Theorem~2]{jaksch2010near}, the value function $u_{l,k}$ computed by \EVI\ at iteration $l$ satisfies: $\max_{s} u_{l,k}(s) - \min_{s} u_{l,k}(s) \leq D$. The convergence criterion of \EVI\ implies
\begin{equation}
|u_{l+1,k}(s) - u_{l,k}(s) - g_k| \leq \frac{1}{\sqrt{t_k}}\,, \qquad  \forall s \in \cS\, .
\label{eq:puterman_convergence}
\end{equation}
Using the Bellman operator on the optimistic MDP, we have:
\als{
u_{l+1,k}(s) 
&= \frac{1}{\ell}\sum_{i=1}^\ell \widetilde\mu_{i,k}((s,\pi^+_{k}(s))[Z_i^r]) + \sum_{s'} \widetilde P_{k}(s'|s,\pi^+_{k}(s)) u_{l,k}(s')\,.
}
Substituting this into (\ref{eq:puterman_convergence}) gives
\begin{align}
\label{eq:eq_EVI_diff}
\Big|\Big( g_k - \frac{1}{\ell}\sum_{i=1}^\ell \widetilde\mu_{i,k}((s,\pi^+_{k}(s))[Z_i^r]) \Big) - \Big(\sum_{s'} \widetilde P_{k}(s'|s,\pi^+_{k}(s))u_{l,k}(s') - u_{l,k}(s)\Big)\Big|
\leq \frac{1}{\sqrt{t_k}}\, , \qquad \forall s\in \cS\, .
\end{align}
Now returning to (\ref{eq:delta_init}), we can write
\begin{align*}
\Delta_k &\leq  \sum_{x} n_k(x) \Big( g_k - \frac{1}{\ell}\sum_{i=1}^\ell  \widetilde\mu_{i,k}(x[Z_i^r]) \Big) + \frac{1}{\ell}\sum_{x} n_k(x) \sum_{i=1}^\ell\big( \widetilde\mu_{i,k}(x[Z_i^r]) - \mu_{i}(x[Z_i^r]) \big) +  \sum_{x} \frac{n_k(x)}{\sqrt{t_k}} \nonumber\, ,
\end{align*}
which, using  (\ref{eq:eq_EVI_diff}), can be simplified as 
\als{
\Delta_k \le \sum_{s} n_k(s,\pi^+_{k}(s)) \Big(\sum_{s'} \widetilde P_{k}(s'|s,\pi^+_{k}(s))u_{l,k}(s') - u_{l,k}(s)\Big) + 
\frac{1}{\ell}\sum_{x} n_k(x) \sum_{i=1}^\ell( \widetilde\mu_{i,k} - \mu_{i})(x[Z_i^r])  +  2\sum_{x} \frac{n_k(x)}{\sqrt{t_k}}
}
Defining ${\bf g}_k = g_k \mathbf 1$, 
$\widetilde{\mathbf{P}}_k := \big(\widetilde P_{k}\big(s'|s,\pi^+_k(s)\big)\big)_{s, s'}$, and $n_{k} := \big(n_k\big(s,\pi^+_k(s)\big)\big)_s$, we  can rewrite the above inequality as:
\begin{align}
\Delta_k &\leq n_{k} (\widetilde{\mathbf{P}}_k - \mathbf{I} ) u_{l,k} + \frac{1}{\ell}\sum_{x} n_k(x) \sum_{i=1}^\ell (\widetilde\mu_{i,k} - \mu_i)(x[Z_i^r]) +  2\sum_{x} \frac{n_k(x)}{\sqrt{t_k}} \, .\nonumber
\end{align}

Similarly to \citep{jaksch2010near}, we define $w_k(s) := u_{l,k}(s) - \tfrac{1}{2}(\min_s u_{l,k}(s) + \max_s u_{l,k}(s))$ for all $s\in \cS$. Then, in view of the fact that $\widetilde{\mathbf{P}}_k$ is row-stochastic, we obtain
\begin{align}
\Delta_k&\leq n_{k} (\widetilde{\mathbf{P}}_k - \mathbf{I} ) w_k + \frac{1}{\ell}\sum_{x} n_k(x) \sum_{i=1}^\ell (\widetilde \mu_{i,k} - \mu_i)(x[Z_i^r]) +  2\sum_{x} \frac{n_k(x)}{\sqrt{t_k}}\, .
\end{align}
The second term in the right-hand side can be upper bounded as follows: $M \in \cM_{k}$ implies that 
\als{
\sum_{i=1}^\ell (\widetilde \mu_{i,k} - \mu_{i})(x[Z_i^r]) &\leq 2\sum_{i=1}^\ell \sqrt{\frac{2\widehat\sigma_{i,k}^2(x[Z_i^r])}{N^r_{i,t}(x[Z_i^r])}\beta_{N^r_{i,k}(x[Z_i^r])}\Big(\frac{\delta}{\ell|\cX[Z_i^r]|}\Big)} + 2\sum_{i=1}^\ell\frac{7}{3N^r_{i,t}(x[Z_i^r])}\beta_{N^r_{i,t}(x[Z_i^r])}\Big(\frac{\delta}{\ell|\cX[Z_i^r]|}\Big)\\
&\leq \sum_{i=1}^\ell \sqrt{\frac{2}{N^r_{i,k}(x[Z_i^r])}\beta_{T}\Big(\frac{\delta}{\ell|\cX[Z_i^r]|}\Big)} + \frac{14}{3}\sum_{i=1}^\ell\frac{1}{N^r_{i,k}(x[Z_i^r])}\beta_{T}\Big(\frac{\delta}{\ell|\cX[Z_i^r]|}\Big)
   \, ,
}
where we have used $\widehat\sigma_{i,k}^2(x[Z_i^r])\le \tfrac{1}{4}$ and $1\leq N^r_{i,k}(x[Z_i^r]) \leq T$ in the last inequality. Furthermore, since  $t_k\ge \max_{i\in [\ell]} N^r_{i,k}(x[Z_i^r])$, we have
\als{
\sum_{x}  \frac{n_k(x)}{\sqrt{t_k}} &\le \sum_{x}  \sum_{i=1}^\ell \frac{n_k(x)}{\sqrt{N^r_{i,k}(x[Z_i^r])}}\, .
}
Putting together, and denoting $\beta_{T,i}:= \beta_{T}\big(\frac{\delta}{\ell|\cX[Z_i^r]|}\big)$, we obtain
\begin{align}
\Delta_k
&\leq n_{k} (\widetilde{\mathbf{P}}_k-\mathbf{I})w_k + 
\frac{\sqrt{2}}{\ell}\sum_{i=1}^\ell  \sqrt{\beta_{T,i}} \sum_{x\in \cX}\frac{n_k(x)}{\sqrt{N_{i,k}^r(x[Z_i^r])}} + \frac{14}{3\ell}\sum_{i=1}^\ell  \beta_{T,i} \sum_{x\in \cX}\frac{n_k(x)}{N_{i,k}^r(x[Z_i^r])} \sk
&\leq n_{k} (\widetilde{\mathbf{P}}_k-\mathbf{I})w_k + 
\frac{\sqrt{2}}{\ell}\sum_{i=1}^\ell  \sqrt{\beta_{T,i}}  \sum_{x\in \cX[Z_i^r]} \frac{\nu_{i,k}^r(x)}{\sqrt{N_{i,k}^r(x[Z_i^r])}} + \frac{14}{3\ell}\sum_{i=1}^\ell  \beta_{T,i}  \sum_{x\in \cX[Z_i^r]} \frac{\nu_{i,k}^r(x)}{N_{i,k}^r(x[Z_i^r])} \, ,
\label{eq:main_delta_minmk}\end{align}
where the last inequality follows from Lemma \ref{lem:factored_count}, stated and proven in Section \ref{sec:supp_lemmas}.

In what follows, we derive an upper bound on $n_{k} (\widetilde{\mathbf{P}}_k-\mathbf{I})w_k$. Similarly to \citep{jaksch2010near}, we consider the following standard decomposition:
\als{
n_k(\widetilde{\mathbf{P}}_k - \mathbf{I}) w_k = \underbrace{n_k (\widetilde{\mathbf{P}}_k - \mathbf{P}_k) w_k}_{L_1(k)} + \underbrace{n_k (\mathbf{P}_k-\mathbf{I})w_k}_{L_2(k)} \, .
}

The following lemmas provide upper bounds on $L_1(k)$ and $L_2(k)$:

\begin{lemma}\label{lem:L_1_k}
Consider a good episode $k$. Then,
\als{
L_1(k) &\leq 3\sum_{i=1}^m \sqrt{\beta'_{T,i}}\sum_{x=(s,a)\in \cX[Z_i^p]} \nu^p_k(x)  D_{i,s}\sqrt{\frac{K_{i,x} - 1}{N^p_{i,k}(x)}}
 + 7\sum_{i=1}^m  DS_i\beta'_{T,i}\sum_{x\in \cX[Z_i^p]}  \frac{\nu_{i,k}^p(x)}{N^p_{i,k}(x)} \, ,
}
where $\beta'_{T,i}:= \beta_T\big(\tfrac{\delta}{2mS_i|\cX[Z_i^p]|}\big)$. 
\end{lemma}

\begin{lemma}\label{lem:L_2_k}
For all $T$, it holds with probability at least $1-\delta$,
\als{
\sum_{k=1}^{K(T)} L_2(k) \bI\{M \in \cM_{k}\} \leq D\sqrt{2T\beta_T(\delta)} + D K(T) \, .
}
\end{lemma}

Applying Lemmas \ref{lem:L_1_k} and \ref{lem:L_2_k}, and summing over all good episodes, we obtain the following bound that holds
 with probability higher than $1-2\delta$, uniformly over all $T\in\Nat$:
\begin{align}
\sum_{k=1}^{K(T)} \Delta_k &\bI\{M \in \cM_{k}\} \leq
\sum_{k=1}^{K(T)} (L_1(k) + L_2(k)) + 
\frac{\sqrt{2}}{\ell}\sum_{i=1}^\ell    \sum_{x\in \cX[Z_i^r]} \frac{\sqrt{\beta_{T,i}}\nu_{i,k}^r(x)}{\sqrt{N_{i,k}^r(x[Z_i^r])}} + \frac{14}{3\ell}\sum_{i=1}^\ell    \sum_{x\in \cX[Z_i^r]} \frac{\beta_{T,i}\nu_{i,k}^r(x)}{N_{i,k}^r(x[Z_i^r])} 
 \nonumber \\
&\leq 3\sum_{i=1}^m \sqrt{\beta'_{T,i}}\sum_{x=(s,a)\in \cX[Z_i^p]}    D_{i,s}\sqrt{K_{i,x} - 1}\frac{\nu^p_{i,k}(x)}{\sqrt{N^p_{i,k}(x)}}
 + 7 \sum_{i=1}^m DS_i\beta'_{T,i}\sum_{x\in \cX[Z_i^p]}  \frac{\nu_{i,k}^p(x)}{N^p_{i,k}(x)} \sk
 &+ \frac{\sqrt{2}}{\ell}\sum_{i=1}^\ell  \sqrt{\beta_{T,i}}  \sum_{x\in \cX[Z_i^r]} \frac{\nu_{i,k}^r(x)}{\sqrt{N_{i,k}^r(x[Z_i^r])}} + \frac{14}{3\ell}\sum_{i=1}^\ell  \beta_{T,i}  \sum_{x\in \cX[Z_i^r]} \frac{\nu_{i,k}^r(x)}{N_{i,k}^r(x[Z_i^r])} \sk
 &+ 2\sum_{i=1}^\ell \sum_{x\in \cX[Z_i^r]} \beta_{T,i}' \frac{n_k(x)}{\sqrt{N_{i,k}^r(x[Z_i^r])}} +  D \sqrt{2T \beta_T(\delta)} + D K(T) \, .
\label{eq:trans_prob_intermed_minmk}
\end{align}

To simplify the above bound, we provide the following lemma:
\begin{lemma}\label{lem:N_k_sequenence_sums}
	Consider a set $\cX'$, and for $x\in \cX'$, let $\nu_k(x)$ (resp.~$N_k(x)$) denote the number of times $x$ is oberserved in episode $k$ (resp.~before episode $k$ starts). We have:
\als{
(i)& \qquad \sum_{x\in \cX'} \sum_{k=1}^{K(T)} \frac{\nu_k(x)\alpha(x)}{\sqrt{N_{k}(x)}}  \leq
	\big( \sqrt{2} + 1 \big) \sqrt{T\sum_{x\in \cX'}\alpha(x)}\, . \\
(ii)& \qquad \sum_{x\in \cX'} \sum_{k=1}^{K(T)} \frac{\nu_k(x)}{N_{k}(x)}  \leq
 2|\cX'|\log\big(\tfrac{T}{|\cX'|}\big) + |\cX'|\, .
}
\end{lemma}

Moreover, following the same steps in the proof of \citep[Proposition~18]{jaksch2010near} to upper bound the number of episodes, we obtain 
\als{
K(T) = \cO\Big(\Big[\sum_{i=1}^m |\cX[Z_i^p]| + \sum_{i=1}^\ell |\cX[Z_i^r]|\Big]\log(T)\Big)\, .
}

Putting everything together, it holds that with probability at least $1-5\delta$,
\als{
\kR(T) &\leq 
8 \sum_{i=1}^m \sqrt{\beta_{T,i}'\sum_{(s,a)\in \cX[Z_i^p]} D^2_{i,s} (K_{i,s,a} - 1) T} + \frac{4}{\ell}\sum_{i=1}^\ell  \sqrt{\beta_{T,i}|\cX[X_i^r]|T} \\
&+ \frac{6}{\ell}\sum_{i=1}^\ell \beta_{T,i}|\cX[Z_i^r]| \log\big(\tfrac{T}{|\cX[Z_i^r]|}\big) + 14\sum_{i=1}^m D S_i\beta_{T,i}'|\cX[Z_i^p]| \log\big(\tfrac{T}{|\cX[Z_i^p]|}\big) \sk
 &+  \Big(D\sqrt{2} + \sqrt{\tfrac{1}{2}}\Big) \sqrt{T\beta_T(\delta)} + 
\cO\Big(D\Big[\sum_{i=1}^m |\cX[Z_i^p]| + \sum_{i=1}^\ell |\cX[Z_i^r]|\Big]\log(T)\Big) \, .
}
Noting that $\beta_T,\beta_{T,i},\beta'_{T,i} = \cO(\log(\log(T)/\delta))$ gives the desired result and completes the proof. 
\ep



\subsection{Proof of Lemma \ref{lem:L_1_k}}
Recall that $L_1(k)  = \sum_{s\in \cS}n_k(s,\pi^+_k(s)) \sum_{y\in \cS} \big(\widetilde P_k -  P_k\big)\big(y|s,\pi^+_k(s)\big) w_k(y)$. 
Fix $s\in \cS$, and introduce $x^k:= (s,\pi^+_k(s))$. We have
\als{
\sum_{y\in \cS} \big(\widetilde P_k -  P_k)\big(y|s,\pi^+_k(s)\big) w_k(y) &\leq \underbrace{\sum_{y\in \cS} \big|\big(\widehat P_k - P_k\big)\big(y|x^k\big)\big| \big|w_k(y)\big|}_{F_1} + \underbrace{\sum_{y\in \cS} \big|\big(\widetilde P_k - \widehat P_k\big)\big(y|x^k\big)\big| \big|w_k(y)\big|}_{F_2}\, .
}

To upper bound $F_1$, recalling the definition $\cK_{i,x}:=\supp(P_i(\cdot|x))$, let us define
$$
H_k(x^k) := \max_{y\in\otimes_{i=1}^m \cK_{i,x^k[Z^p_i]}} |w_k(y)|\, , 
$$
Now an application of Lemma \ref{lem:factored_deviation} gives:
\als{
F_1 &\leq  H_k(x^k)\sum_{i=1}^m \sqrt{\frac{2\beta_{T,i}'}{N^p_{i,k}(x^k[Z^p_i])}} \sum_{y[i]\in \cS_i} \sqrt{P_{i}(1-P_i)(y[i]|x^k[Z^p_i])}  + \sum_{i=1}^m \frac{ DS_i\beta_{T,i}'}{2N^p_{i,k}(x^k[Z^p_i])}\, .
}
where we used that $\max_{y\in \cS} |w_k(y)|\leq \tfrac{D}{2}$ since $M\in \cM_k$ (following a similar argument as in \citep{jaksch2010near}), and where we used the shorthand $\beta_{T,i}':=\beta_T(\tfrac{\delta}{2mS_i|\cX[Z_i^p]|})$. 
 
By Cauchy-Schwarz, we get
\als{
\sum_{y[i]\in \cS_i} &\sqrt{P_i(1-P_i)(y|x^k[Z^p_i])} =  \sum_{y[i]\in \cK_{i,x^k[Z^p_i]}} \sqrt{P_i(1-P_i)(y[i]|x^k[Z^p_i])} \\
&\leq  \sqrt{  \sum_{y[i]\in \cK_{i,x^k[Z^p_i]}  } P_i(y[i]|x^k[Z_i^p])} \sqrt{\sum_{y[i]\in \cK_{i,x^k[Z^p_i]}  } (1-P_i)(y[i]|x^k[Z_i])} \\
&\leq  \sqrt{K_{i,x^k[Z^p_i]} -1} \, ,
}
so that
\als{
F_1 &\leq H_k(x^k)\sum_{i=1}^m \sqrt{\frac{2\beta_{T,i}'}{N^p_{i,k}(x^k[Z^p_i])}} \sqrt{K_{i,x^k[Z^p_i]} -1}  + \sum_{i=1}^m \frac{DS_i\beta_{T,i}'}{2N^p_{i,k}(x^k[Z^p_i])} \, .
}

To upper bound $F_2$, we will need the following lemma:

\begin{lemma}[{\citep{bourel2020tightening}}]
	\label{lem:SqrtVar_pqBern}
	Consider $x$ and $y$ satisfying $|x - y| \leq \sqrt{2y(1-y)\zeta} + \zeta/3$. Then,
\begin{align*}
\sqrt{y(1 - y)} &\leq \sqrt{x(1- x)} + 2.4\sqrt{\zeta} \, .
	\end{align*}
\end{lemma}

Applying this lemma twice, we have that for any $y$ and $x$, if $|\widetilde P_{i,k} - \widehat P_{i,k}|(y|x) \leq \sqrt{2\widetilde P_{i,k}(1-\widetilde P_{i,k})(y|x)\zeta} + \zeta'/3$, where $\zeta$ and $\zeta'$ come from the definition of $C_{t,\delta,i}(x,y)$, then
\als{
|\widetilde P_{i,k} - \widehat P_{i,k}|(y|x) \leq \sqrt{2P_i(1-P_i)(y|x)\zeta} + 4\zeta'. 
}
Hence, an application of Lemma  \ref{lem:factored_deviation} gives 
\als{
F_2 &\leq H_k(x^k)\sum_{i=1}^m \sqrt{\frac{2\beta_{T,i}'}{N^p_{i,k}(x^k[Z^p_i])}} \sqrt{K_{i,x^k[Z^p_i]} -1}  + 6 \sum_{i=1}^m \frac{DS_i\beta_{T,i}'}{N^p_{i,k}(x^k[Z^p_i])} \, .
}

Putting together, we get
\als{
L_1(k) &\leq 3\sum_{x\in \cX} n_k(x)H_k(x)\sum_{i=1}^m \sqrt{\frac{\beta_{T,i}'}{N^p_{i,k}(x[Z^p_i])}} \sqrt{K_{i,x[Z^p_i]} -1} 
 + 7 \sum_{x\in \cX} \nu^p_k(x) \sum_{i=1}^m \frac{DS_i\beta_{T,i}'}{N^p_{i,k}(x[Z^p_i])}  \, .
}

To control the right-hand side, we further show that given $x\in \cX$,
$$
H_k(x) \leq \max_{u=(s,a): u[Z_i^p]=x}\max_{y\in \cL_s} |w_k(y)| \leq D_{i,s[Z_i^p]}, \quad \forall i\in [m].
$$
where $\cL_{s} := \otimes_{i=1}^m(\cup_{a\in \cA[Z^p_i]} \cK_{i, s[Z_i^p], a})$

The first inequality holds by the definition of $H_k$. To verify the second claim,  we note that similarly to \citep{jaksch2010near}, we can combine all MDPs in $\cM_k$ to form a single MDP $\widetilde\cM_k$ with continuous action space $\cA'$. In this extended MDP, in each state $s\in \cS$, and for each $a\in \cA$, there is an action in $\cA'$ with mean $\tilde\mu(s,a)$ and transition $\widetilde P(\cdot|s,a)$ satisfying the definition of the set of plausible MDPs. Similarly to the arguments in \citep{jaksch2010near}, we recall that $u_{l,k}(s)$ amounts to the total expected $l$-step reward of an optimal non-stationary $l$-step policy starting in state $s$ on the MDP $\widetilde\cM_k$ with extended action set. Recall that we are in a case where $M\in \cM_k$. This implies that the local diameter of factor $i$ and state $s$ of this extended MDP is at most $D_{i,s[Z_i^p]}$, since the actions of the true MDP are contained in the continuous action set of the extended MDP $\widetilde\cM_k$. 
Let 
\als{
\cB_i:= \Big\{y\in \cS: y[i]\in \cup_{a'\in \cA[Z_i^p]} \cK_{i,s[Z_i^p], a'}\, \hbox{ and }  \, y[j]\in \cS_j, i\neq j\Big\}.
}
Now, if there were states $s_1,s_2\in \cB_i$ with $u_{l,k}(s_1) - u_{l,k}(s_2) >  D_{i,s[Z_i^p]}$, then an improved value
for $u_{l,k}(s_1)$ could be achieved by the following non-stationary policy: First follow a policy which moves from $s_1$ to $s_2$ most quickly, which takes at most $D_{i,s[Z_i^p]}$ steps on average. Then follow the optimal $l$-step policy for $s_2$. We thus have $u_{l,k}(s_1) \geq u_{l,k}(s_2)- D_{i,s[Z_i^p]}$, since at most $D_{i,s[Z_i^p]}$ rewards of the policy for $s_2$ are missed. This is a contradiction, and so the claim follows. 
	

Hence,
\als{
L_1(k) &\leq 3\sum_{i=1}^m \sum_{x\in \cX} n_k(x) D_{i,s[Z_i^p]} \sqrt{\frac{\beta_{T,i}'}{N^p_{i,k}(x[Z^p_i])}} \sqrt{(K_{i,x[Z^p_i]} - 1) }
 + 7 \sum_{i=1}^m \sum_{x\in \cX}  n_k(x)\frac{DS_i\beta_{T,i}'}{N^p_{i,k}(x[Z^p_i])} \\
 &\leq 3\sum_{i=1}^m \sum_{x=(s,a)\in \cX[Z_i^p]} \nu^p_{i,k}(x)  D_{i,s}\sqrt{\frac{\beta_{T,i}' (K_{i,x} - 1)}{N^p_{i,k}(x)}}
 + 7 \sum_{i=1}^m DS_i\beta_{T,i}' \sum_{x\in \cX[Z_i^p]}  \frac{\nu_{i,k}^p(x)}{N^p_{i,k}(x)} 
 \, ,
}
where the last inequality follows from Lemma \ref{lem:factored_count}. 
\ep

\subsection{Proof of Lemma \ref{lem:L_2_k}}
The proof follows similar steps as in the proof of \citep[Theorem~2]{jaksch2010near}. Here, we collect all necessary arguments for completeness. Let us define the sequence $(X_t)_{t\geq 1}$ with $X_t := (P(\cdot|s_t,a_t) - \mathbf e_{s_{t+1}})w_{k(t)}\bI\{M \in \cM_{k(t)}\}$, for all $t$, where $k(t)$ denotes the episode containing step $t$. For any good $k$ (i.e., $M \in \cM_{k}$), as established in the proof of \citep[Theorem~2]{jaksch2010near}, it holds that:
	\als{
		L_2(k) &= \nu_k (\mathbf{P}_k-\mathbf{I})w_k  = \sum_{t=t_k}^{t_{k+1} -1} (P(\cdot|s_t,a_t) - \mathbf{e}_{s_t}) w_k 
				= \sum_{t=t_k}^{t_{k+1} -1} X_t + w_k(s_{t+1}) - w_k(s_t) \leq \sum_{t=t_k}^{t_{k+1} -1} X_t + D \, ,
	}
	so that $\sum_{k=1}^{K(T)} L_2(k) \leq \sum_{t=1}^T
	X_t +K(T)D$. 
Moreover, if $k$ is a good episode, as shown in \citep[Theorem~2]{jaksch2010near}, $|X_t| \leq \|P(\cdot|s_t,a_t) - \mathbf e_{s_{t+1}}\|_1\frac{D}{2} \leq  D$. Further, $\mathbb{E}[X_t|s_1, a_1, \dots, s_t, a_t] = 0$, so that $(X_t)_t$ is martingale difference sequence with $|X_t|\le D$. Therefore, applying Lemma  \ref{lem:bounded_peeling} gives:
	\beqan
	\mathbb P\Big(\exists  T: \sum_{t=1}^T X_t  \geq D\sqrt{2T\beta_T(\delta)} \Big)\leq \delta\,.
	\eeqan
	Putting this together with the above bound on $\sum_{k=1}^{K(T)} L_2(k)$ gives the desired result. 
\ep
	
%

\subsection{Other Supporting Lemmas}
\label{sec:supp_lemmas}
\subsubsection{Proof of Lemma \ref{lem:N_k_sequenence_sums}} 
Observe that for any sequence of numbers $z_1, z_2, \dots, z_n$ with $0 \leq z_k \leq Z_{k-1} := \max\{1, \sum_{i=1}^{k-1}z_i\}$, it holds
\als{
\sum_{k=1}^n 	\frac{z_k}{\sqrt{Z_{k-1}}} 	\leq 	\big(\sqrt{2} + 1\big) 	\sqrt{Z_n} \quad \hbox{and}\quad \sum_{k=1}^n \frac{z_k}{Z_{k-1}} \leq 2\log(Z_n) + 1\, .
}
We refer to \citep[Lemma~19]{jaksch2010near} and \citep[Lemma~24]{talebi2018variance} for proof of these facts. The assertion of the lemma then easily follows by these facts and using Jensen's inequality. 
\ep

%

\subsubsection{Lemma \ref{lem:factored_count} and Its Proof}

\begin{lemma}\label{lem:factored_count}
Let $Z_1,\ldots,Z_m\subseteq [n]$, and for $x\in \cX$, let $\nu_k(x)$ denote the local counts for some episode $k$. Then, for any $i$ and any positive function $\alpha:\cX[Z_i] \to \RR$, we have: 
\als{
\sum_i \sum_{x\in \cX} \nu_k(x) \alpha(x[Z_i]) \leq  \sum_i \sum_{x'\in \cX[Z_i]} \nu_k(x')\alpha(x')\, .
}
\end{lemma}

\bp
We have:
\als{
\sum_i \sum_{x\in \cX}\nu_k(x) &\alpha(x[Z_i]) = \sum_i \sum_{x\in \cX} \sum_{t=t_k}^{t_{k+1}-1}  \indic{x_{t'} = x} \alpha(x[Z_i])\\
&= \sum_i \sum_{x'\in \cX[Z_i]}\sum_{x''\in \cX[[n]\setminus Z_i]}  \sum_{t=t_k}^{t_{k+1}-1}  \indic{x_{t'}[[n]\setminus Z_i] = x''} \indic{x_{t'}[Z_i] = x} \alpha(x[Z_i])\\
&= \sum_i \sum_{x'\in \cX[Z_i]} \alpha(x')  \sum_{t=t_k}^{t_{k+1}-1} \indic{x_{t'}[Z_i] = x}\underbrace{\sum_{x''\in \cX[[n]\setminus Z_i]}   \indic{x_{t'}[[n]\setminus Z_i] = x''}}_{\leq 1}
\\
&\leq \sum_i \sum_{x'\in \cX[Z_i]} \nu_k(x')\alpha(x')\, . 
}
\ep

%



\subsubsection{Proof of Lemma \ref{lem:SqrtVar_pqBern}}
The proof is provided in \citep{bourel2020tightening} and is presented here for completeness. 
By Taylor's expansion, we have
	\als{
		y(1-y) &= x (1- x) + (1 - 2x)(y - x) -(y - x)^2 \\
		&= x(1- x) + (1 - x - y)(y - x) \\
		&\leq
		x (1-x) + |1 - x - y |\left(\sqrt{2y (1-y)\zeta} + \tfrac{1}{3}\zeta\right) \sk
		&\leq
		x (1-x) + \sqrt{2 y(1- y)\zeta} + \tfrac{1}{3}\zeta
		\, .
	}
	Using the fact that $a\le b\sqrt{a}+c$ implies $a\le b^2 + b\sqrt{c} + c$ for nonnegative numbers $a,b$, and $c$, we get
	\begin{align}
	y(1-y)
	&\leq
	x (1-x)  + \tfrac{1}{3}\zeta + \sqrt{2\zeta\left(x (1-x)  + \tfrac{1}{3}\zeta\right)} + 2\zeta \sk
	&\leq
	x (1-x)  + \sqrt{2\zeta x (1-x)} + 3.15\zeta  \sk
	&= \left(\sqrt{x (1-x)}  + \sqrt{\tfrac{1}{2}\zeta}\right)^2 + 2.65\zeta\, ,
	\end{align}
	where we have used $\sqrt{a+b}\le \sqrt{a} + \sqrt{b}$ valid for all $a,b\ge 0$. Taking square-root from both sides and using the latter inequality give the desired result:
	\begin{align*}
	\sqrt{y(1 - y)} &\leq
	\sqrt{x(1- x)} + \sqrt{\tfrac{1}{2}\zeta} + \sqrt{2.65\zeta} \leq \sqrt{x(1- x)} + 2.4\sqrt{\zeta} \, .
	\end{align*}
\ep

\section{\uppercase{Regret Analysis: Cartesian Products}}
\label{app:regret_proof_Cartesian}

\subsection{Proof of Lemma \ref{lem:VI_Cartesian_prod}}

\textbf{Lemma 2 (Restated)} 
\emph{
Consider VI with $u_0(s) = 0$, and for each $n\ge 0$, $u_{n+1}(s)=\max_{a\in \cA}\big\{m^{-1}\mu(s,a) + \sum_{y\in \cS} P(y|s,a) u_n(y)\big\}$. Then, for all $n$, $u_{n}(s) = m^{-1}\sum_{i=1}^m u^{(i)}_{n}(s[i])$, where $(u^{(i)}_{n})_{n\geq 0}$ is a sequence of VI on MDP $i$, that is $u^{(i)}_0(x)=0$ and $u^{(i)}_{n+1}(x) = \max_{a\in \cA_i}\big\{\mu_i(x,a) +  \sum_{y\in \cS_i} P_i(y|x,a) u^{(i)}_{n}(y)\big\}$ for all $x\in \cS_i$ and $n\geq 0$. 
}

\bp
We prove the lemma by induction on $n$. 
Consider $u_0(s) = 0$. For $n=1$, we have:
\als{
u_1(s) &= \max_{a\in \cA} \frac{\mu(s,a)}{m} = \max_{a[1],\ldots,a[m]} \sum_{i} \frac{\mu_i(s[i],a[i])}{m} = \frac{1}{m}\sum_i \max_{a[i]\in \cA_i} \mu_i(s[i],a[i]) = \frac{1}{m}\sum_i u_1^{(i)} \, .
}
Now assume that the induction hypothesis is correct for $n>1$, that is, $u_n(s) = m^{-1}\sum_{i=1}^k u^{(i)}_{n}(s[i])$. We would like to show that $u_{n+1}(s) = m^{-1}\sum_{i=1}^k u^{(i)}_{n+1}(s[i])$. We have
\als{
u_{n+1}(s) &= \max_{a\in \cA}\bigg\{\mu(s,a)/m + \sum_{y\in \cS} P(y|s,a) u_n(y)\bigg\} \\
&= \max_{a[1],\ldots,a[m]}\bigg\{\frac{1}{m}\sum_{i} \mu_i(s[i],a[i]) + \sum_{y_1\in \cS^{(1)}}\ldots\sum_{y_m\in \cS_m} \prod_{i=1}^m P_i(y_i|s[i],a[i]) \sum_{j=1}^m \frac{u^{(j)}_{n}(y_j)}{m} \bigg\} \, .
}
We have
\als{
\sum_{y_1\in \cS_1}\!\ldots\!\sum_{y_m\in \cS_{m}} \prod_{i=1}^m P_i\big(y_i|s[i],a[i]\big) \sum_{j=1}^m u^{(j)}_{n}(y_j) &= \sum_{j=1}^m \sum_{y_j\in \cS_j} P_j(y_j|s[j],a[j])u^{(j)}_{n}(y_j)\!\sum_{y \in \otimes_{i\neq j} \cS_i} \prod_{i\neq j} P_i(y_i|s[i],a[i]) \\
&= \sum_{j=1}^m \sum_{y_j\in \cS_j} P_j(y_j|s[j],a[j])u^{(j)}_{n}(y_j) \,.
}
Hence, 
\als{
mu_{n+1}(s) &= \max_{a[1],\ldots,a[m]}\bigg\{\sum_{i} \mu_i(s[i],a[i]) + \sum_{i=1}^m \sum_{y_i\in \cS_i} P_i(y_i|s[i],a[i])u^{(i)}_{n}(y_i) \bigg\} \\
&= \sum_{i=1}^m \max_{a[i]\in \cA_i}\bigg\{\mu_i(s[i],a[i]) + \sum_{y_i\in \cS_i} P_i(y_i|s[i],a[i])u^{(i)}_{n}(y_i) \bigg\}\\
&= \sum_{i=1}^m u_{n+1}^{(i)}(s[i]) \, ,
}
thus concluding the lemma. 
\ep

\subsection{Proof of Theorem \ref{thm:regret_Alg1_Cartesian}}

Without loss of generality, we assume $\cG_r = \cG_p$ (and in particular, $\ell = m$), and further assume that $(Z_{i}^r)_{i\in [\ell]}$ (and so, $(Z_{i}^p)_{i\in [m]}$) forms a partition of $[n]$ -- Hence, with no loss of generality, the FMDP is assumed to be the product of $m$ base MDPs. The proof can be straightforwardly extended to the case where $\cG_r \neq \cG_p$, at the expense of more complicated and tedious notations. 

As $\cG_r = \cG_p$, in what follows we omit the dependence of scope sets or various quantities on $r$ and $p$. 
We first make the following observation, which follows by straightforward calculations: $g^\star = \tfrac{1}{m}\sum_{i=1}^m g_i^\star$, where $g^\star_i$ denotes the optimal gain in $M_i$. 
We have 
\als{
\kR(T) &= Tg^\star - \sum_{t=1}^T\frac{1}{m}\sum_{i=1}^m r_{t}[i] = \frac{1}{m}\sum_{i=1}^m \big(Tg^\star_i - \sum_{t=1}^T r_{t}[i]\big) \, .
}
Following similar steps as in the proof of Theorem \ref{thm:regret_Alg1}, we obtain that with probability at least $1-\delta$, 
\als{
\kR(T) \leq \sum_{k=1}^{K(T)} \Delta_k + \sqrt{T\beta_T(\delta)/2}\, ,
}
where $\Delta_k$ is defined similarly to the  proof of Theorem \ref{thm:regret_Alg1}. Now consider a good episode $k$. The corresponding optimistic policy  $\pi^+_{k}$ and $\widetilde M_{k}$ satisfy (as in the  proof of Theorem \ref{thm:regret_Alg1}): $g_k := g_{\pi^+_{k}}^{\widetilde M_{k}} \geq g^\star - \frac{1}{\sqrt{t_k}}$. In particular,  by construction $\widetilde M_k\in \mathbb M_{\cG(M)}$, and thus, $g_k = \tfrac{1}{m}\sum_{i=1}^m g_{k}^{(i)}$, where $g_{k}^{(i)}$ denotes the gain of the restriction of policy $\pi^+_{k}$ to the base MDP $i$ in $\widetilde M_k$. 
We therefore have:
\begin{align*}
\Delta_k  &\leq \sum_{x\in \cX}  \frac{n_k(x)}{m}\sum_{i=1}^m \big(g_{k}^{(i)} - \mu_i(x[Z_i])\big) + \sum_{x\in \cX} \frac{n_k(x)}{\sqrt{t_k}} 
\, .
\end{align*}
Leveraging the arguments in the proof of  Theorem \ref{thm:regret_Alg1}, and applying Lemma \ref{lem:VI_Cartesian_prod},\footnote{We stress that Lemma \ref{lem:VI_Cartesian_prod} applies to \EVI\ as well, as the inner maximization of \EVI\ is guaranteed to return a factored transition function.} we observe that the value function $u_{l,k}$ computed by \EVI\ at iteration $l$ satisfies:
\als{
u_{l+1,k}(s) &= \frac{1}{m}\sum_{i=1}^m \widetilde\mu_{i,k}((s,\pi^+_{k}(s))[Z_i]) + \sum_{s'} \widetilde P_k(s'|s,\pi^+_{k}(s)) u_{l,k}(s')\\
&=\frac{1}{m} \sum_{i=1}^m \Big(\widetilde\mu_{i,k}((s,\pi^+_{k}(s))[Z_i]) + \sum_{y_i\in \cS_i} \widetilde P_{i,k}(y_i|(s,\pi^+_{k}(s))[Z_i]) u_{l,k}^{(i)}(y_i) \Big)
\, .
} 
Now the stopping criterion of \EVI\ implies:
\begin{equation*}
\Big| \sum_{i=1}^m \Big( g_{i,k} - \widetilde\mu_{i,k}((s,\pi^+_{k}(s))[Z_i]) - \sum_{y_i\in \cS_i} \widetilde P_{i,k}(y_i|(s,\pi^+_{k}(s))[Z_i])u_{i,k}^{(l)}(y_i) + u_{i,k}^{(l)}(s[i])\Big)\Big|
\leq \frac{m}{\sqrt{t_k}}\, , \qquad \forall s\in \cS\, ,
\end{equation*}
which, after following similar steps as in the proof of Theorem \ref{thm:regret_Alg1}, gives
\begin{align}
\Delta_k&\leq  \frac{1}{m}\sum_{x} n_k(x)  \sum_{i=1}^m \Big(g_{k}^{(i)} -  \widetilde\mu_{i,k}(x[Z_i]) \Big) + \frac{1}{m}\sum_{x} n_k(x) \sum_{i=1}^m\Big( \widetilde\mu_{i,k}(x[Z_i]) - \mu_{i}(x[Z_i]) \Big) +  2\sum_{x} \frac{n_k(x)}{\sqrt{t_k}} \nonumber\\
&\leq  \frac{1}{m}\sum_{x} n_k(x)  \sum_{i=1}^m \Big(g_{k}^{(i)} -  \widetilde\mu_{i,k}(x[Z_i]) \Big) + \frac{1}{m}\sum_{x} n_k(x) \sum_{i=1}^m\Big( \widetilde\mu_{i,k}(x[Z_i]) - \mu_{i}(x[Z_i]) \Big) \sk
& + 2\sum_{x}  \sum_{i=1}^m \frac{n_k(x)}{\sqrt{N_{i,k}(x[Z_i])}}
\nonumber\\
&\leq    \frac{1}{m}\sum_{i=1}^m \sum_{x\in \cX[Z_i]} \nu_{i,k}(x) \Big(g_{k}^{(i)} -  \widetilde\mu_{i,k}(x) \Big) + \frac{1}{m}\sum_x \sum_{i=1}^m n_k(x)\frac{2\beta'_{T,i}}{\sqrt{N_{i,k}(x[Z_i])}}  +  2\sum_{i=1}^m \sum_{x\in \cX[Z_i]} \frac{\nu_{i,k}(x)}{\sqrt{N_{i,k}(x)}} \nonumber\\
&\leq \frac{1}{m}\sum_{i=1}^m \nu_{i,k} (\widetilde{\mathbf{P}}_{i,k} - \mathbf{I} ) u_{l,k}^{(i)} + \frac{1}{m}\sum_{i=1}^m \beta'_{T,i}\sum_{x\in \cX[Z_i]} \frac{\nu_{i,k}(x)}{\sqrt{N_{i,k}(x)}} +  2\sum_{i=1}^m \sum_{x\in \cX[Z_i]} \frac{\nu_{i,k}(x)}{\sqrt{N_{i,k}(x)}}  \, .\nonumber
\end{align}
Here, we used the fact that $t_k\ge \max_{i\in [m]} N_{i,k}(x[Z_i])$, and applied Lemma \ref{lem:factored_count}. 

Now defining for each $i$, $w_{i,k}(s) := u_{l,k}^{(i)}(s) - \tfrac{1}{2}(\min_s u_{l,k}^{(i)}(s) + \max_s u_{l,k}^{(i)}(s))$ for all $s\in \cS_i$, we arrive at
$\Delta_k\leq \frac{1}{m}\sum_{i=1}^m \widetilde \Delta_{i,k}$ with 
\als{
\widetilde \Delta_{i,k} := \nu_{i,k} (\widetilde{\mathbf{P}}_{i,k} - \mathbf{I} ) u_{l,k}^{(i)} + \beta'_{T,i} \sum_{x\in \cX[Z_i]} \frac{\nu_{i,k}(x)}{\sqrt{N_{i,k}(x)}} +  2 \sum_{x\in \cX[Z_i]} \frac{\nu_{i,k}(x)}{\sqrt{N_{i,k}(x)}} \, . 
}
In other words, the regret is upper bounded by a quantity, which only depends on the properties of MDP $M_i$. Following exact same arguments as in the rest of proof of Theorem \ref{thm:regret_Alg1} (or similarly, those in the proof of Thoerem 1 in \citep{bourel2020tightening}) gives the desired result. 
\ep

\section{\uppercase{Details of The Example for The Factored Diameter}}
\label{app:diameter}

\begin{figure}[!t]
\begin{center}
	\includegraphics[scale=1]{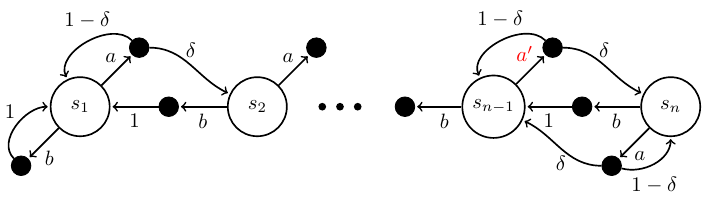}
    \caption{Global vs.~Factored Diameter}
 	\label{fig:diameter}
\end{center}
\end{figure}

Let us first consider a single agent. In this case, by definition, action $a'$ is absent. The form of the transition function here allows us to derive closed-form expressions for the notions of diameters. In particular, the (global) diameter is $D=\frac{n-1}{\delta}$, as it takes $\frac{n-1}{\delta}$ steps in expectations to reach $s_n$ from $s_1$ (this is the worst-case shortest path between any pair of states). Moreover, the local diameter \citep{bourel2020tightening} is upper bounded by $2/\delta$: For any state $s$, this is the worst-case shortest path between any pair of states taken among the possible next-states of $s$. 

Now we consider the case of $2$ agents, where each agent independently interacts with an instance of the $n$-state MDP shown in Figure \ref{fig:diameter}. 
Each agent $i\in\{1,2\}$ occupies a state in $\cS_i=\{s_1,s_2,\ldots,s_n\}$ with $n>2$, and the transition function $P_i$ is defined according to the MDP shown in the figure.
In each state $s\neq s_{n-1}$, each agent has access to two actions $a$ and $b$. Only when both agents are \emph{simultaneously} in $s_{n-1}$, they have access to an extra action $a'$, which causes each agent to stochastically (but independently) transit to a high-reward state $s_n$ --- For instance, this could be relevant in scenarios where cooperation yields higher rewards. 
This scenario can be modeled using an FMDP as follows. The state-space is $\cS=\cS_1\times \cS_2$ and the action-space is state-dependent: in each state $s\neq (s_{n-1},s_{n-1})$, $\cA_s=\{a,b\}\times \{a,b\}$, whereas in $s=(s_{n-1},s_{n-1})$, $\cA_s =\{a',b\} \times \{a',b\}$. 

The state space of the full MDP (modeling the interactions of both agents with the environment) is denoted by $\cS$: 
$$
\cS = \Big\{s=(s[1],s[2]): s[1]\in \cS_1, s[2]\in \cS_2\Big\}
$$ 
It thus has $|\cS_1|\times |\cS_2| = n^2$ states. 
For example, the possible transitions at the state $s$, where $s[1]=s_1$ and $s[2]=s_1$ would be as follows. Under action $(a,a)$: the next is $(s_1,s_1)$ w.p. $(1-\delta)^2$, or $(s_2,s_1)$ w.p. $\delta(1-\delta)$, or $(s_1,s_2)$ w.p.~$\delta(1-\delta)$, or $(s_2,s_2)$ w.p.~$\delta^2$. Under action $(b,b)$, the next state is $(s_1,s_1)$ w.p.~$1$. Under action $(b,a)$ the next state is $(s_1,s_2)$ w.p.~$\delta$ or $(s_1,s_1)$ w.p.~$1-\delta$. Finally, under action $(a,b)$, the next state is $(s_2,s_1)$ w.p.~$\delta$ or $(s_1,s_1)$ w.p.~$1-\delta$. 

In the full MDP, it is easy to verify that $D=\big(\tfrac{n-1}{\delta}\big)^2$. But the local diameter of the full MDP (which corresponds to the factored diameter of the FMDP) is at most $\tfrac{4}{\delta^2}$. Recalling the definition of the factored diameter, it is straightforward to see that, for $s=(s_i,s_j)$, we have $\cL_{s} = \{s_{(i-1)\vee 1},s_{i},s_{(i+1)\wedge n}\}\times\{s_{(j-1)\vee 1},s_{j},s_{(j+1)\wedge n}\}$ --- We use shorthands $a\vee b\!=\!\max\{a,b\}$ and $a\wedge b\!=\!\min\{a,b\}$. So one can verify that for any two states $u,v\in \cL_s$, it takes at most  $\tfrac{4}{\delta^2}$ steps in expectation to reach $u$ starting from $v$.

\end{document}